\def\year{2020}\relax
\documentclass[letterpaper]{article} 
\usepackage{aaai20}  
\usepackage{times}  
\usepackage{helvet} 
\usepackage{courier}  
\usepackage[hyphens]{url}  
\usepackage{graphicx} 
\usepackage{graphicx}  
\usepackage{subfiles}
\usepackage{mathtools}  
\usepackage[ruled,vlined,linesnumbered]{algorithm2e}
\usepackage{amsmath}
\usepackage{amssymb}
\usepackage{amsthm}
\usepackage{enumerate}
\usepackage{threeparttable}
\usepackage{booktabs} 
\graphicspath{{imgs/}}
\usepackage{subcaption}

\input{required.tex}

\title{Efficient Projection-Free Online Methods \\ with Stochastic Recursive Gradient}
\author{Jiahao Xie\textsuperscript{1}, Zebang Shen\textsuperscript{2}, Chao Zhang\textsuperscript{1,3}, Boyu Wang\textsuperscript{4}, Hui Qian\textsuperscript{1}\\
\textsuperscript{1}College of Computer Science and Technology, Zhejiang University\\
\textsuperscript{2}University of Pennsylvania\\
\textsuperscript{3}Tencent AI Lab\\
\textsuperscript{4}University of Western Ontario\\
xiejh@zju.edu.cn, zebang@seas.upenn.edu, zczju@zju.edu.cn, bwang@csd.uwo.ca, qianhui@zju.edu.cn
}
\begin{document}

\maketitle

\begin{abstract}
This paper focuses on projection-free methods for solving smooth Online Convex Optimization (OCO) problems.
Existing projection-free methods either achieve suboptimal regret bounds or have high per-iteration computational costs.
To fill this gap, two efficient projection-free online methods called ORGFW and MORGFW are proposed for solving stochastic and adversarial OCO problems, respectively.
By employing a recursive gradient estimator, our methods achieve optimal regret bounds (up to a logarithmic factor) while possessing low per-iteration computational costs.
Experimental results demonstrate the efficiency of the proposed methods compared to state-of-the-arts.
\end{abstract}

\section{Introduction}  \label{section_introduction}

We consider the following smooth Online Convex Optimization (OCO) problem.
In each round $t = 1, \ldots, T$, a learner chooses a decision $\xB_t$ from a compact convex set $\CM \subseteq \RBB^d$.
Then a smooth convex loss function $f_t$ is revealed and the learner suffers the loss $f_t(\xB_t)$.
We consider both adversarial and stochastic settings.
In the \textit{adversarial setting}, the sequence of functions $\{f_t\}_{t=1}^T$ can be arbitrary (possibly adversarial), while in the \textit{stochastic setting}, the loss functions are sampled i.i.d. from some fixed distribution $f_t \sim \DM$.
The target of the learner is to produce a sequence of decisions $\{\xB_t\}$ that minimizes the regret, which is the cumulative loss suffered by the learner compared to that of the best fixed decision in hindsight, i.e.,
$$
    {\textstyle\sum}_{t=1}^T f_t(\xB_t) - {\textstyle\sum}_{t=1}^T f_t(\xB^*),
$$
where $\xB^* \in \mathrm{argmin}_{\xB \in \CM} \sum_{t=1}^T f_t(\xB)$ in the adversarial setting and $\xB^* \in \mathrm{argmin}_{\xB \in \CM} \EBB_{f_t \sim \DM}[f_t(\xB)]$ in the stochastic setting, respectively.

This model captures a wide range of real-world applications in which data points arrive sequentially,
e.g., online recommendation systems, online spam email filtering, online prediction in financial markets, online portfolio selection, to name a few~\cite{hazan2016introduction,hoi2018online,cesa2006prediction,agarwal2006algorithms}.

Existing methods for solving OCO problems can be divided into two categories: (i) projection-based methods~\cite{zinkevich2003online,shalev2007primal,xiao2010dual,duchi2011adaptive,cutkosky2017stochastic}, and (ii) projection-free methods~\cite{hazan2012projection,hazan2016introduction,lafond2015online,chen2018projection}.
For high-dimensional applications with complicated constraints (e.g., low rank matrix completion~\cite{chandrasekaran2009sparse}, network routing~\cite{hazan2016introduction}, and structural SVMs~\cite{julien2013block}), the projection operation can be computationally expensive or even intractable, rendering projection-based methods impractical.
In contrast, projection-free methods such as Frank-Wolfe-type methods~\cite{hazan2012projection,hazan2016introduction} only require to solve linear optimization problems over the constraint set, which is usually simpler than projection.
Thus projection-free methods have attracted considerable attention in recent years.

\begin{table*}[t]
	\begin{threeparttable}
	\caption{Comparison of projection-free online methods.
	The fourth column shows the per-iteration computational cost in average.
	The fifth column indicates whether a method uses stochastic gradients or exact (full) gradients of $f_t$'s.}
	\label{table_related_algorithms}
	\begin{tabular}{@{}p{\textwidth}@{}}
	\centering
	\begin{tabular}{c|c|c|c|c|c}
		\toprule
		Algorithm       & Setting  & Regret & Per-round cost     & Stochastic gradients    & Guarantee \\
		\midrule
		OFW   & adversarial, cvx. & $\OM(T^{3/4})$  & $\OM(T)$ & No  & deterministic \\
		Regularized OFW   & adversarial, cvx., smooth & $\OM(T^{3/4})$  & $\OM(1)$  & No  & deterministic \\
		Meta-Frank-Wolfe   & adversarial, cvx., smooth & $\OM(\sqrt{T})$  & $\OM(T^{3/2})$  & Yes  & in expectation  \\
		\textbf{MORGFW (this work)}   & adversarial, cvx., smooth & $\tilde{\OM}(\sqrt{T})$ &  $\OM(T)$  & Yes  & w.h.p.  \\
		\midrule
		OFW   & stoch., cvx., smooth & $\tilde{\OM}(\sqrt{T})$ & $\OM(T)$ & No  & w.h.p.   \\
		OSFW   & stoch., cvx., smooth & $\OM(T^{2/3})$ & $\OM(1)$ & Yes  & in expectation  \\
		\textbf{ORGFW (this work)}   & stoch., cvx., smooth & $\tilde{\OM}(\sqrt{T})$ & $\OM(1)$  & Yes  & w.h.p. \\
		\bottomrule
	\end{tabular}
	\end{tabular}
	\begin{tablenotes}
		\footnotesize
		\item
		OFW~\cite{hazan2012projection},
		Regularized OFW~\cite{hazan2016introduction},
		OSFW and Meta-Frank-Wolfe~\cite{chen2018projection}
	\end{tablenotes}
	\end{threeparttable}
\end{table*}

However, existing projection-free online methods suffer from a trade-off between regret and computational complexity.
The seminal work, the Online Frank-Wolfe (OFW) method~\cite{hazan2012projection}, achieves an $\OM(T^{3/4})$ regret in the adversarial setting.
Besides, in the stochastic setting,
OFW achieves a nearly optimal $\tilde{\OM}(\sqrt{T})$ regret\footnotemark with high probability.
\footnotetext{It is known that the optimal regret bound for general OCO problems is $\OM(\sqrt{T})$ (see, e.g.,~\cite{hazan2016introduction}). We call $\tilde{\OM}(\sqrt{T})$ nearly optimal, where $\tilde{\OM}()$ suppresses a poly-logarithmic factor. }
For both settings, the per-iteration computational cost of OFW ($\OM(T)$ on average) is considerably high.
The Regularized OFW method~\cite{hazan2016introduction} improves the per-iteration computational cost to $\OM(1)$ in the adversarial setting while remaining the same $\OM(T^{3/4})$ regret bound as OFW.
Additionally, OFW and Regularized OFW require to access exact gradients of $f_t$'s, which can be computationally prohibitive in online applications with large-scale streaming data where a large batch of data arrives in each round~\cite{dekel2012optimal}.
To tackle this problem, \citeauthor{chen2018projection}~\shortcite{chen2018projection} propose two methods called Meta-Frank-Wolfe and OSFW, which use stochastic gradient estimates, for the adversarial and stochastic settings, respectively.
Meta-Frank-Wolfe requires $\OM(T^{3/2})$ stochastic gradient evaluations in each round, although it achieves the optimal regret bound ($\OM(\sqrt{T})$ in expectation) in the adversarial setting.
OSFW achieves a suboptimal $\OM(T^{2/3})$ regret bound (in expectation) in the stochastic setting.
To the best of our knowledge, none of existing projection-free online methods has both the optimal (or nearly optimal) regret bound and a low computational cost at the same time.

To bridge this gap, we propose two novel projection-free methods, Online stochastic Recursive Gradient-based Frank-Wolfe (ORGFW) and Meta-ORGFW (MORGFW), for OCO problems in the stochastic and adversarial settings, respectively.
Both methods achieve nearly optimal regret bounds with high probability while having low computational costs.
To achieve this goal, we utilize a recursive variance reduction technique to reduce noise in stochastic gradients without bringing much extra computation.
Then, we develop a new analysis technique based on martingale concentration inequalities to bound the gradient approximation error to a desired accuracy, which allows us to derive the optimal regret bound.
Note that a similar variance reduction technique has been adopted by~\cite{cutkosky2019momentum} for solving unconstrained nonconvex stochastic optimization.
While they focus on finding an approximate stationary point, we aim at producing a sequence of decisions that has low regret.
Our contributions are listed as follows.
\begin{itemize}
    \item
    We show that ORGFW achieves a nearly optimal $\tilde{\OM}(\sqrt{T})$ regret bound for smooth OCO problems in the stochastic setting.
    To the best of our knowledge, this is the first projection-free online method that has both a nearly optimal regret bound and an $\OM(1)$ per-iteration computational cost in such setting.
    \item For smooth OCO problems in the adversarial setting, MORGFW achieves a nearly optimal $\tilde{\OM}(\sqrt{T})$ regret bound.
    This method only requires $\OM(T)$ stochastic gradient evaluations in each round, improving upon the $\OM(T^{3/2})$ cost of Meta-Frank-Wolfe~\cite{chen2018projection}.
    \item Compared to the regret bounds in~\cite{chen2018projection}, which hold in expectation, our results hold with high probability and therefore rule out the possibility that the regret has high variance.
    To establish high-probability regret bounds, we propose a new analysis technique by utilizing a martingale concentration inequality to bound the gradient approximation error in high probability. This technique can be of independent interests for establishing high-probability bounds for other online methods.
\end{itemize}
A summary of our results and previous ones is provided in Table~\ref{table_related_algorithms}.
In addition to regret bounds for online learning, we also prove the convergence of ORGFW for solving both convex and nonconvex stochastic optimization problems.
Our experimental results demonstrate the advantages of the proposed methods over existing projection-free methods.

\section{Related Work}  \label{section_related_work}

\textbf{Online projection-free methods.}
The classical Frank-Wolfe (FW) method (a.k.a. conditional gradient descent) is introduced by~\cite{frank1956algorithm} for solving offline optimization problems.
Starting with~\cite{hazan2008sparse}, Frank-Wolfe has regained a lot of popularity because it has the advantages of projection-free, norm-free, and sparse iterates~\cite{bubeck2015convex}.
\citeauthor{hazan2012projection}~\shortcite{hazan2012projection} propose the first online Frank-Wolfe method called OFW, which requires to evaluate the gradient of the cumulative loss function $\sum_{\tau = 1}^t f_t$ at the $t$-th iteration and thus has a high computational cost in general.
\citeauthor{lafond2015online}~\shortcite{lafond2015online} propose Online Away-step Frank-Wolfe (OAW), which incorporates the away step technique~\cite{guelat1986some} into OFW.
They show that both OFW and OAW achieve logarithmic regrets for OCO problems in the stochastic setting if the loss functions are strongly convex and smooth and the constraint set satisfies additional assumptions.
Besides, they also prove that these two methods find a stationary point of a nonconvex stochastic optimization problem.
\citeauthor{hazan2016introduction}~\shortcite{hazan2016introduction} proposes a method called Regularized OFW, which leverages a regularization technique and only requires to evaluate one gradient of $f_t$ at the $t$-th iteration.
\citeauthor{zhang2017projection}~\shortcite{zhang2017projection} extends Regularized OFW to distributed online learning for solving OCO problems with large-scale streaming data.
Another direction to solve large-scale OCO problems is to reduce the computational cost by using stochastic gradient estimates instead of exact gradients of $f_t$'s, which is studied in~\cite{chen2018projection}.

\textbf{Variance reduction.}
Variance Reduction (VR) techniques are originally proposed to reduce variance in gradient estimation for stochastic gradient methods~\cite{johnson2013accelerating,defazio2014saga,nguyen2017sarah,fang2018spider,zhou2018stochastic,nguyen2018inexact}.
Several stochastic projection-free VR methods have been proposed  for solving offline optimization problems~\cite{hazan2016variance,reddi2016stochastic,mokhtari2018stochastic,shen2019complexities,yurtsever2019conditional}.
These VR methods cannot directly apply to OCO problems since OCO problems are fundamentally different from offline ones.
Recently, \citeauthor{chen2018projection}~\shortcite{chen2018projection} propose the first projection-free VR method for OCO.

\section{Notation and Preliminaries} \label{section_preliminaries}

\textbf{Notation.} We use bold lowercase symbols (e.g., $\xB$) to denote vectors and bold uppercase symbols (e.g., $\AB$) to denote matrices.
The entry in the $i$-th row and $j$-th column of a matrix $\AB$ is denoted by $[\AB]_{ij}$.
Throughout this paper, we use $\|\xB\|$ to denote the standard Euclidean norm of a vector $\xB$.

We consider both the \textit{adversarial setting} and the \textit{stochastic setting} of online convex optimization problems.
For these two settings, the definitions of regret are slightly different.
In the adversarial setting, the regret is defined as
\begin{equation}  \label{eq_regret_convex_adversarial}
    \mathcal{R}_T := \sum_{t=1}^T f_t(\xB_t) -\min_{\xB \in \CM} \sum_{t=1}^T f_t(\xB).
\end{equation}
In the stochastic setting, the regret is defined as
\begin{equation} \label{eq_regret_convex_stochastic}
    \mathcal{SR}_T := \sum_{t=1}^T \big( f_t(\xB_t) - f_t(\xB^*) \big),
\end{equation}
where $\xB^* \in \mathrm{argmin}_{\xB \in \CM} \fbar(\xB) := \EBB_{f_t \sim \DM} [f_t(\xB)]$.
We note that the OCO problem in the stochastic setting is closely related to but different from the \textit{stochastic optimization} problem~\cite{birge1997introduction}.
In OCO, the goal is to produce a sequence of decision variables that has low regret and the learner must properly respond to the environment (i.e., updating the decision variable) as soon as new data arrive~\cite{dekel2012optimal}.
In stochastic optimization, however, we aim to find an approximate minimizer of the loss function and the performance of a method is measured by the convergence rate.
Compared to the OCO problem, stochastic optimization focuses on the quality of the final output of a method instead of the sequence of iterates produced over the course of optimization.

\section{Online Stochastic Recursive Gradient-Based Frank-Wolfe}  \label{section_algorithm}

In this section, we present our projection-free methods for solving OCO problems.
We first introduce the Online stochastic Recursive Gradient-based Frank-Wolfe (ORGFW) method, which uses a stochastic recursive gradient estimator, for the stochastic setting.
Based on ORGFW, we introduce the Meta-ORGFW (MORGFW) method for the more challenging adversarial setting.

\begin{algorithm}[t]
\SetKwInOut{Input}{Input}
\SetKwInOut{Return}{Return}
\SetAlgoLined
\Input{parameters $\{\rho_t\}_{t=1}^T$, $\{\eta_t\}_{t=1}^T$, and initial point $\xB_0 = \xB_1 \in \CM$}

\For{$t = 1, 2, \ldots, T$}
{
    Play $\xB_t$, then receive $f_t(\xB_t)$ and stochastic gradients $\nabla F_t(\xB_{t-1}, \xi_t)$ and $\nabla F_t(\xB_{t}, \xi_t)$; \

    \uIf{$t = 1$}{
        $\dB_t \leftarrow \nabla F_t(\xB_t, \xi_t)$; \
    }\Else{
        $\dB_{t} \leftarrow \nabla F_t(\xB_{t}, \xi_t) + (1 - \rho_{t}) \big( \dB_{t-1} - \nabla F_t(\xB_{t-1}, \xi_t) \big)$; \
    }
    $\vB_t \leftarrow \underset{\vB \in \CM}{\mathrm{argmin}} \langle \dB_t, \vB \rangle$; \

    $\xB_{t+1} \leftarrow  \xB_t + \eta_t (\vB_t - \xB_t)$; \
}
\caption{ORGFW}
\label{algorithm_ORGFW}
\end{algorithm}

\subsection{Algorithm in the Stochastic Setting}
Now we present ORGFW, which is detailed in Algorithm~\ref{algorithm_ORGFW}.
In each round $t = 1, \ldots, T$, ORGFW plays $\xB_t$ and receives the loss $f_t(\xB_t)$ as well as stochastic gradients $\nabla F_t(\xB_{t-1}, \xi_t)$ and $\nabla F_t(\xB_{t}, \xi_t)$, where $\xi_t$ is a random variable following some distribution $\PM_t$ such that $\EBB_{\xi_t \sim \PM_t} [ \nabla F_t(\xB, \xi_t) ] = \nabla f_t(\xB)$.
For example, if $f_t$ has a finite-sum structure of the form $f_t(\xB) = \frac{1}{n_t}\sum_{i=1}^{n_t} f_{t, i}(\xB)$, which occurs in online problems with large-scale streaming data, one can let $\PM_t$ be the uniform distribution over $\{1, \ldots, n_t\}$ and $\nabla F_t(\cdot, \xi_t) = \nabla f_{t, \xi_t}(\cdot)$.
In line 6 of ORGFW, we estimate the gradient of $\fbar$ using a stochastic recursive estimator
\begin{equation} \label{eq_storm}
\begin{aligned}[b]
    \dB_{t} =
        \nabla F_t(\xB_{t}, \xi_t) + (1 - \rho_{t}) \big( \dB_{t-1} - \nabla F_t(\xB_{t-1}, \xi_t) \big)
\end{aligned}
\end{equation}
where $\dB_1 = \nabla F_1(\xB_1, \xi_1)$ and $\rho_t$ is a parameter to be determined later.
If the exact gradient $\nabla f_t(\xB)$ can be efficiently computed, one can directly replace $\nabla F_t(\xB, \xi_t)$ with $\nabla f_t(\xB)$.
After updating $\dB_t$, ORGFW finds a solution $\vB_t$ to the linear optimization problem $\mathrm{argmin}_{\vB \in \CM} \langle \dB_t, \vB \rangle$ and updates $\xB_{t+1}$ along the direction $\vB_t - \xB_t$, where the step size $\eta_t$ will be determined later.

The recursive estimator~\eqref{eq_storm} is inspired by~\cite{cutkosky2019momentum} in which a similar estimator is devised for solving unconstrained nonconvex stochastic optimization problems.
One difference between~\eqref{eq_storm} and the estimator in~\cite{cutkosky2019momentum} is that in~\eqref{eq_storm}, the stochastic gradients $\nabla F_t(\cdot, \xi_t)$ in different rounds are sampled from different distributions, while in their estimator, all the stochasticity comes from the same distribution $\PM$.
More importantly, our analysis is fundamentally different from theirs.
In our analysis, we explicitly show that the gradient approximation error converges to zero at a sublinear rate w.h.p., which is critical to analyzing regret bounds in high probability.
In contrast, \citeauthor{cutkosky2019momentum}~\shortcite{cutkosky2019momentum} do not explicitly analyze the convergence property of the approximation error but instead construct a Lyapunov function to derive convergence analysis of their method.

\subsection{Algorithm in the Adversarial Setting}
Inspired by the Meta-Frank-Wolfe method~\cite{chen2018projection}, we use the recursive estimator~\eqref{eq_storm} to develop a meta algorithm called MORGFW for OCO problems in the adversarial setting.
Note that MORGFW is a general framework that relies on the outputs of base Online Linear Optimization (OLO)\footnote{
Online linear optimization is a special case of online convex optimization in which the loss functions are linear.} algorithms.
The MORGFW method is detailed in Algorithm~\ref{algorithm_meta_ORGFW}.
In each round $t = 1, \ldots, T$, it simulates a $K$-step Frank-Wolfe subroutine using stochastic gradients of $f_t$ and OLO algorithms $\EM^{(1)}, \ldots, \EM^{(k)}$.
We refer to $\EM^{(k)}$ for $k \in \{1, \ldots, K\}$ as the base algorithms.
Typical algorithms for OLO include Follow the Perturbed Leader~\cite{kalai2005efficient}, Online Gradient Descent~\cite{zinkevich2003online}, Regularized-Follow-The-Leader~\cite{shalev2007primal}, etc.
From line 2 to line 6 in MORGFW, we sequentially take $K$ Frank-Wolfe-type update steps in which the update direction $\vB_t^{(k)}$ is produced by the base algorithm $\EM^{(k)}$.
We then take the final iterate $\xB_t^{(K+1)}$ as the prediction in the $t$-th round and receive the loss function $f_t$ as well as the stochastic gradient oracle.
From line 8 to line 16 in MORGFW, we sequentially compute $\dB_t^{(k)}$ using the recursive estimator to approximate $\nabla f_t(\xB_t^{(k)})$ for $k = 1, \dots, K$ and feedback the linear loss $\langle \dB_t^{(k)}, \vB_t^{(k)} \rangle$ to $\EM^{(k)}$.
If the exact gradient $\nabla f_t$ can be efficiently computed, one can directly replace $\dB_t$ with $\nabla f_t(\xB_t)$.

\begin{algorithm}[t]
\SetKwInOut{Input}{Input}
\SetKwInOut{Return}{Return}
\SetAlgoLined
\Input{Parameters $T$, $K$, $\{\eta_k\}_{k=1}^{K}$, $\{\rho_k\}_{k=1}^{K}$, base algorithms $\EM^{(1)}, \ldots, \EM^{(K)}$, initial point $\xB_1 \in \CM$}

\For{$t = 1, 2, \ldots, T$}
{
    Initialize $\xB_t^{(1)} = \xB_1$; \

    \For{$k = 1, 2, \ldots, K$}
    {
        $\vB_t^{(k)} \leftarrow$ output of $\EM^{(k)}$ in round $t-1$; \

        $\xB_t^{(k+1)} \leftarrow (1 - \eta_k)\xB_t^{(k)} + \eta_k \vB_t^{(k)}$; \
    }

    Play $\xB_t = \xB_t^{(K+1)}$ and receives $f_t(\xB_t)$ and stochastic gradient oracle $\nabla F_t(\cdot, \cdot)$; \

    \For{$k = 1, 2, \ldots, K$}
    {
        Sample $\xi_t^{(k)} \sim \PM_t$; \

        \uIf{$k = 1$}{
            $\dB_t^{(k)} \leftarrow \nabla F_t(\xB_t^{(k)}, \xi_t^{(k)})$; \
        }\Else{
            $\dB_{t}^{(k)} \leftarrow \nabla F_t(\xB_t^{(k)}, \xi_t^{(k)}) + (1 - \rho^{k}) \big( \dB_{t}^{(k-1)} - \nabla F_t(\xB_{t}^{(k-1)}, \xi_t^{(k)}) \big)$; \
        }

        Feedback $\langle \vB_t^{(k)}, \dB_t^{(k)} \rangle$ to $\EM^{(k)}$;
    }

}
\caption{MORGFW}
\label{algorithm_meta_ORGFW}
\end{algorithm}

\section{Regret Analysis}  \label{section_analysis}

In this section, we analyze the regret bounds of the proposed methods.
As a byproduct, we also derive convergence guarantee of ORGFW for convex and nonconvex stochastic optimization problems, respectively.
All missing proofs are deferred to the \textbf{Appendix} in the supplementary material due to the limit of space.
To begin with, we make the following two common assumptions on the constraint set $\CM$ and stochastic gradients of $f_t$'s, respectively.
\begin{assumption} \label{assumption_constraint}
    The compact convex set $\CM \subseteq \RBB^d$ has diameter $D$, i.e., $\forall \xB, \yB \in \CM$,
    $$
        \| \xB - \yB \| \le D.
    $$
\end{assumption}

\begin{assumption}  \label{assumption_adversarial_smooth}  \label{assumption_stoch_smooth}
    The stochastic gradient $\nabla F_t(\xB, \xi_t)$ is unbiased (i.e., $\EBB_{\xi_t}[\nabla F_t(\xB, \xi_t)] = \nabla f_t(\xB)$) and is $L$-Lipschitz continuous over the constraint set $\CM$, i.e.,
    $$
        \| \nabla F_t(\xB, \xi_t) - \nabla F_t(\yB, \xi_t) \| \le L \|\xB - \yB\|, \ \forall \xB, \yB \in \CM.
    $$
\end{assumption}
Assumption~\ref{assumption_stoch_smooth} immediately implies that $f_t$ is differentiable and has $L$-Lipschitz-continuous gradients.

\subsection{Analysis of ORGFW}
In the stochastic online setting, we denote the expected loss function as $\fbar = \EBB_{f_t \sim \DM}[f_t]$.
In order to obtain high probability results, the following common assumption is required.

\begin{assumption}  \label{assumption_stoch}
    We assume the following
    \begin{enumerate}[a]
        \item  \label{assumption_stoch_gradient}
        The distance between the stochastic gradient $\nabla F_t(\xB, \xi_t)$ and the exact gradient is bounded over the constraint set $\CM$, i.e., for any $\xB \in \CM$, $t \in \{1, \ldots, T\}$, there exists $\sigma^2 < \infty$ such that with probability $1$,
        $$
            \| \nabla F_t(\xB, \xi_t) - \nabla \fbar(\xB)\|^2 \le \sigma^2.
        $$

        \item \label{assumption_stoch_value}
        The difference of $f_t(\xB)$ and $\fbar(\xB)$ is bounded over the constraint set $\CM$, i.e., $\forall \xB \in \CM$, $t \in \{1, \ldots, T\}$, there exists $M^2 < \infty$ such that with probability $1$,
        $$
            |f_t(\xB) - \fbar(\xB)|^2 \le M^2.
        $$
    \end{enumerate}
\end{assumption}

In our proofs, we develop a new analysis technique to show that the norm of the gradient estimation error
$\epsilonB_t := \dB_t - \nabla \fbar(\xB_t)$ converges to zero rapidly w.h.p.
The main idea of our analysis technique is summarized in the following and the detailed proof is deferred to Appendix~\ref{section_proof_key_lemma}.
First, we reformulate $\epsilonB_t$ as the sum of a martingale difference sequence $\{\zetaB_{t, \tau}\}_{\tau=1}^{t}$ w.r.t. a filtration $\{\FM_{\tau}\}_{\tau=0}^t$, i.e.,
$
\epsilonB_t = \sum_{\tau = 1}^t \zetaB_{t, \tau},
$
where $\EBB[\zetaB_{t, \tau} | \FM_{\tau-1}] = \zeroB$ and $\FM_{\tau-1}$ is the $\sigma$-filed generate by $\{f_1, \xi_1, \ldots, f_{\tau-1}, \xi_{\tau-1}\}$.
By showing that $\|\zetaB_{t, \tau}\| \le c_{t, \tau}$ for some constant $c_{t, \tau}$, one can relate the error $\|\epsilonB_t\|$ to the quantity $q_t := \sum_{\tau = 1}^t c_{t, \tau}^2$ via an Azuma-Hoeffding-type concentration inequality (see Proposition~\ref{proposition_concentration} in the Appendix).
With carefully chosen $\{\rho_t\}_{t=1}^T$ and $\{\eta_t\}_{t=1}^T$, the quantity $q_t$ can be shown to converge to zero at a sublinear rate by induction.
As a result, $\|\epsilonB_t\|$ converges to zero at a sublinear rate w.h.p. as stated in the following lemma.

\begin{lemma} \label{lemma_gradient_error_bound_whp}
    Consider ORGFW with $\eta_t = \rho_t = 1/(t + 1)^{\alpha}$ for some $\alpha \in (0, 1]$.
    If Assumptions~\ref{assumption_constraint}, \ref{assumption_stoch_smooth}, and~\ref{assumption_stoch}.\ref{assumption_stoch_gradient} are satisfied,
    for any $t \ge 1$ and $\delta_0 \in (0, 1)$, we have w.p. at least $1 - \delta_0$,
    $$
        \| \epsilonB_t \| \le 2 (2 L D + \frac{3^{\alpha} \sigma}{3^{\alpha} - 1})(t + 1)^{- \alpha / 2} \sqrt{2 \mathrm{log}(4 / \delta_0)}.
    $$
\end{lemma}
Lemma~\ref{lemma_gradient_error_bound_whp} shows that the gradient approximation error $\| \epsilonB_t \|$ converges to zero at a fast sublinear rate $\tilde{\OM}(1/t^{\alpha / 2})$ w.h.p. if $\eta_t = \rho_t = 1/(t+1)^{\alpha}$ for any $\alpha \in (0, 1]$.
This result is critical to the regret analysis of our methods.

Now we are ready to present the first main theorem.
\begin{theorem} \label{theorem_convex}
    (Regret Bound w.h.p. in the Stochastic Setting)
    Consider ORGFW with $\eta_t = \rho_t = 1/(t + 1)$.
    If $\fbar$ is convex and Assumptions~\ref{assumption_constraint}-\ref{assumption_stoch} are satisfied,
    then w.p. at least $1 - \delta$ for any $\delta \in (0, 1)$,
    $$
    \begin{aligned}
        \SM \RM_T &\le (\mathrm{log} T + 1) \big( \fbar(\xB_1) - \fbar(\xB^*) \big)
            + \frac{ L D^2 (\mathrm{log}T + 1)^2 }{2} \\
            & \ \ \ \ + (16 L D^2 + 16 \sigma D + 4 M) \sqrt{2 T \mathrm{log}(8 T / \delta)},
    \end{aligned}
    $$
    where $\SM \RM_T$ is defined in~\eqref{eq_regret_convex_stochastic}.
\end{theorem}
Theorem~\ref{theorem_convex} shows that ORGFW achieves a nearly optimal $\tilde{\OM}(\sqrt{T})$ regret bound w.h.p. for OCO problems in the stochastic setting under mild assumptions.
As a byproduct, we provide convergence guarantee of ORGFW for convex stochastic optimization in the following corollary.
\begin{corollary} \label{corollary_convex_convergence}
    (Convergence rate for Convex Stochastic Optimization)
    Assume that $\fbar$ is convex and Assumptions~\ref{assumption_constraint}, \ref{assumption_stoch_smooth}, and~\ref{assumption_stoch}.\ref{assumption_stoch_gradient} are satisfied.
    If we run ORGFW with $\rho_t = \eta_t = 1/(t+1)$ and let $\hat{\xB} = \frac{1}{T} \sum_{t=1}^T \xB_t$, we have, with probability at least $1 - \delta$ for any $\delta \in (0, 1)$,
    $$
    \begin{aligned}
        & \fbar(\hat{\xB}) - \fbar(\xB^*)
        \le \frac{\mathrm{log}T + 1}{T} \big( \fbar(\xB_1) - \fbar(\xB^*) \big)  \\
        & \ \ \ \ + \frac{ L D^2 (\mathrm{log}T + 1)^2 }{2 T}
            + 16 (L D^2 + \sigma D) \frac{\sqrt{2 \mathrm{log}(4 T / \delta)}}{\sqrt{T}}.
    \end{aligned}
    $$
\end{corollary}
Corollary~\ref{corollary_convex_convergence} shows that ORGFW achieves a convergence rate of $\tilde{\OM}(1 / \sqrt{T})$ w.h.p. for convex stochastic optimization problems.
In other words, ORGFW needs $\tilde{\OM}(1/\epsilon^2)$ stochastic gradient evaluations to find a solution $\hat{\xB}$ such that $\fbar(\hat{\xB}) - \fbar(\xB^*) \le \epsilon$, which matches the state-of-the-art result~\cite{lan2016conditional,yurtsever2019conditional}.

Similarly, one can also prove that ORGFW finds an approximate stationary point of a nonconvex stochastic optimization problem.
A point $\xB \in \CM$ is called an $\epsilon$-approximate stationary point if it satisfies the condition
\begin{equation} \label{eq_FW_gap}
    \GM(\xB) := \max_{\uB \in \CM} \langle \nabla \fbar(\xB), \xB - \uB \rangle \le \epsilon,
\end{equation}
where the non-negative quantity $\GM(\xB)$ is known as the Frank-Wolfe gap.
The following proposition establishes the convergence rate of ORGFW to a stationary point.
\begin{proposition} \label{proposition_nonconvex_convergence}
    (Convergence rate for Nonconvex Stochastic Optimization)
    Assume that Assumptions~\ref{assumption_constraint}, \ref{assumption_stoch_smooth}, and~\ref{assumption_stoch}.\ref{assumption_stoch_gradient} are satisfied.
    If we run ORGFW with $\rho_t = \eta_t = 1 / (t + 1)^{2/3}$, we have w.p. at least $1 - \delta$,

    $$
    \begin{aligned}
        \min_{1 \le t \le T} \GM(\xB_t)
        &\le \frac{2 (\fbar(\xB_1) - \fbar(\xB^*))}{T^{1/3}} +
            \frac{4 \mathrm{log}(T + 1)}{T^{1/3}} \Big( L D^2 \\
        & \ \ \ \ + (2 L D^2 + 3 \sigma^2) \sqrt{2 \mathrm{log}(4 T / \delta)} \Big).
    \end{aligned}
    $$
\end{proposition}
Proposition~\ref{proposition_nonconvex_convergence} shows that ORGFW converges to a stationary point at a rate of $\tilde{\OM}(1 / T^{1/3})$.
In other words, ORGFW finds an $\epsilon$-approximate stationary point in $\tilde{\OM}(1 / \epsilon^3)$ stochastic gradient evaluations.
This result outperforms the $\tilde{\OM}(1 / \epsilon^4)$ bound of OFW and OAW~\cite{lafond2015online} and matches the state-of-art result~\cite{shen2019complexities,yurtsever2019conditional}.

\subsection{Analysis of MORGFW}
In the adversarial online setting, we make the following assumption which is analogous to Assumption~\ref{assumption_stoch}.\ref{assumption_stoch_gradient}.

\begin{assumption}  \label{assumption_adversarial}  \label{assumption_adversarial_gradient}
    The distance between the stochastic gradient $\nabla F_t(\xB, \xi_t)$ and the exact gradient $f_t(\xB)$ is bounded over the constraint set $\CM$ (with probability $1$), i.e., $\forall \xB \in \CM$, $t \in \{1, \ldots, T\}$, there exists $\hat{\sigma}^2 < \infty$ such that
    $$
        \| \nabla F_t(\xB, \xi_t) - \nabla f_t(\xB)\|^2 \le \hat{\sigma}^2.
    $$
\end{assumption}

In the following theorem, we establish the regret bound of MORGFW in the adversarial setting.
\begin{theorem} \label{theorem_adversarial}
    (Regret Bound w.h.p. in the Adversarial Setting)
    Consider MORGFW with $K = T$, $\rho_k = \eta_k = 1/(k+1)$.
    Suppose that each of the base algorithms $\EM^{(1)}, \ldots, \EM^{(K)}$ has a regret $\RM_T^{\EM}$.
    If each $f_t$ is convex and Assumptions~\ref{assumption_constraint}, \ref{assumption_adversarial_smooth}, and~\ref{assumption_adversarial} are satisfied, then w.p. at least $1 - \delta$,
    $$
    \begin{aligned}
        \RM_T
        &\le 16 (L D^2 + \hat{\sigma} D) \sqrt{2 T \mathrm{log}(4 T^2 / \delta)} \\
        & \ \ \ \ + 2 L D^2 \mathrm{log}(T+1) + Q + \RM_T^{\EM},
    \end{aligned}
    $$
    where $\RM_T$ is defined in~\eqref{eq_regret_convex_adversarial}, $Q = \max_{1 \le t \le T} \{f_t(\xB_1) - f_t(\xB^*)\}$, and $\xB^* \in \mathrm{argmin}_{\xB \in \CM} \sum_{t=1}^T f_t(\xB)$.
\end{theorem}

Theorem~\ref{theorem_adversarial} shows that the regret of MORGFW is bounded from above by $\tilde{\OM}(\sqrt{T}) + \RM_T^{\EM}$ w.h.p., where $\RM_T^{\EM}$ is the regret of the base algorithms $\EM^{(1)}, \dots, \EM^{(K)}$.
It remains to choose proper base algorithms for OLO.
A suitable choice is Follow the Perturbed Leader which is a projection-free method with $\RM_T^{\EM} = \OM(\sqrt{T})$ for OLO~\cite{kalai2005efficient,cohen2015following}.
Thus, by choosing Follow the Perturbed Leader as the base algorithm $\EM^{(k)}$ in MORGFW, we achieve a nearly optimal $\tilde{\OM}(\sqrt{T})$ regret.

We remark that in Theorem~\ref{theorem_adversarial}, the parameter $K$ is dependent on the time horizon $T$.
Thus, MORGFW requires prior knowledge of the time horizon.
Nevertheless, this issue can be easily solved by the doubling trick~\cite[Section 2.3.1]{shalev2012online}.
Indicated by~\cite{shalev2012online}, the regret bound only increases by a constant multiplicative factor if we adopt this trick.

\section{Experiments}  \label{section_experiments}

To validate the theoretical results in the previous section, we first conduct numerical experiments on an OCO problem, i.e., online multiclass logistic regression.
To further show the efficiency of the proposed methods, we also conduct experiments on an offline nonconvex optimization problem -- training a constrained one-hidden-layer neural network.
We use two well-known multiclass datasets: MNIST\footnote{\url{http://yann.lecun.com/exdb/mnist/}} and CIFAR10\footnote{\url{https://www.cs.toronto.edu/~kriz/cifar.html}}.
Detailed information of these datasets are listed in Table~\ref{table_datasets}.
For all compared methods, we choose hyperparameters via grid search and simply set the initial point to $\zeroB$.
Besides, we repeat the random methods for $6$ trails and report the average result.

\begin{figure}[t]
\begin{subfigure}{.23\textwidth}
  \centering
  \includegraphics[width=\linewidth]{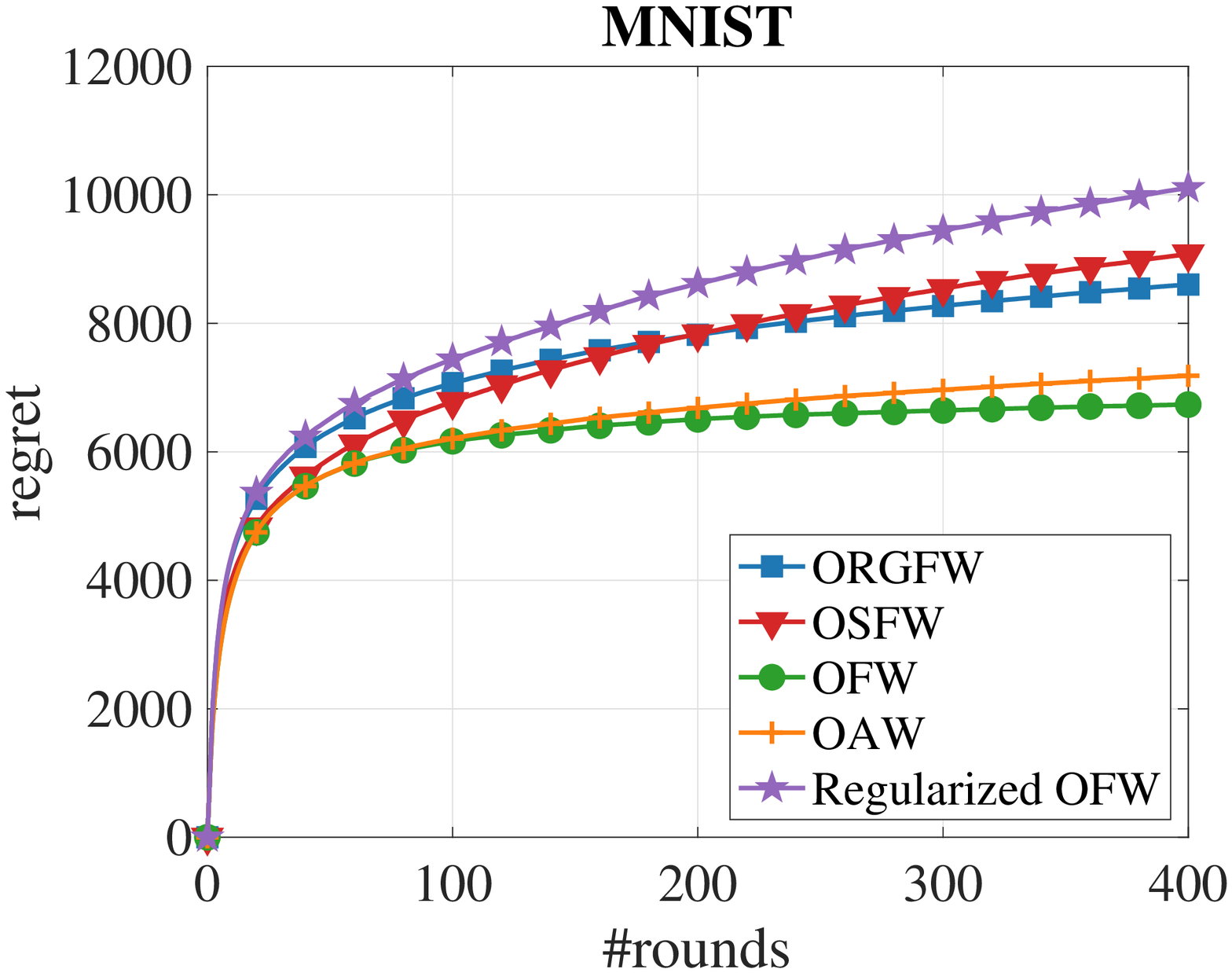}
\end{subfigure}
\hfill
\begin{subfigure}{.23\textwidth}
  \centering
  \includegraphics[width=\linewidth]{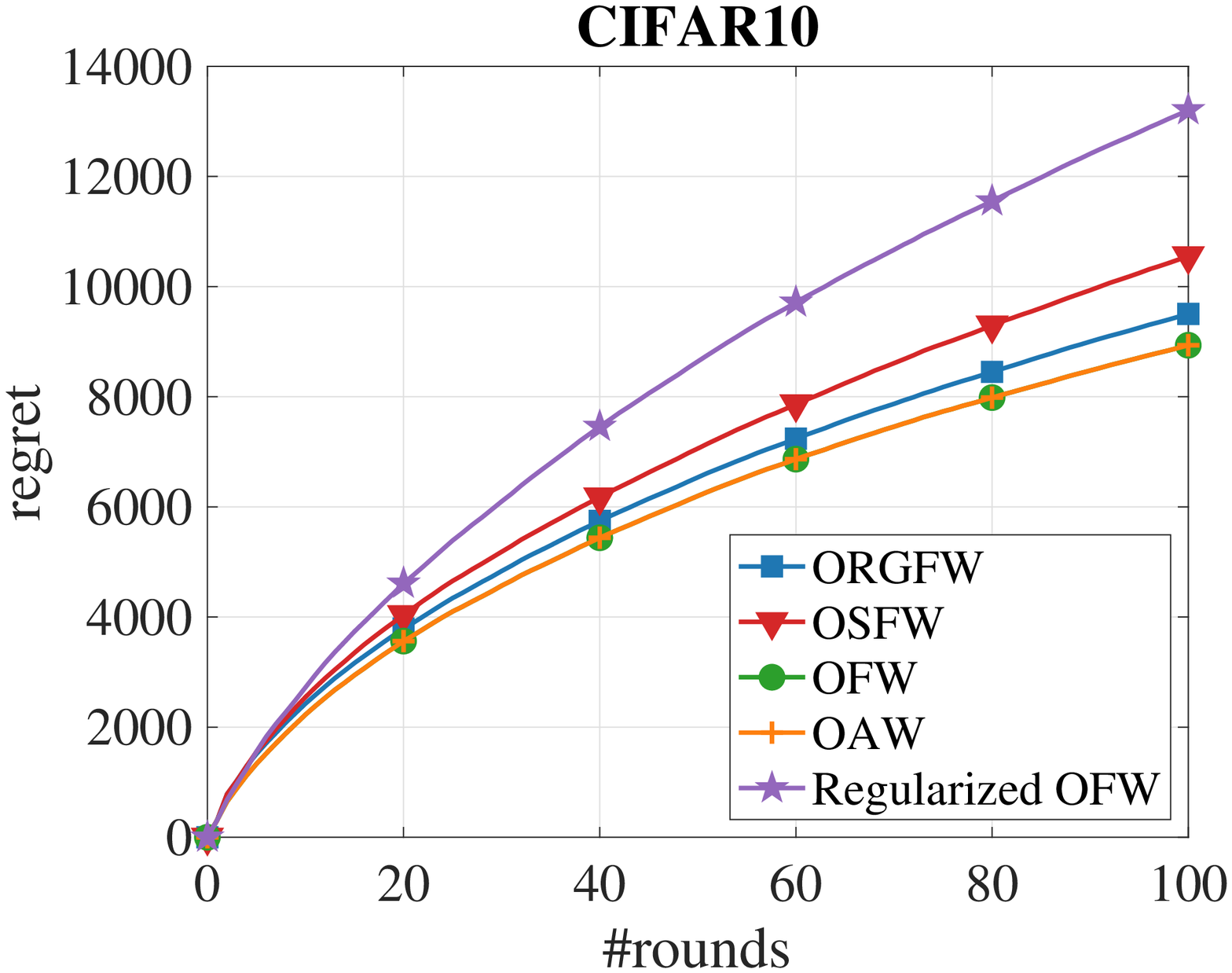}
\end{subfigure}
\\
\begin{subfigure}{.23\textwidth}
  \centering
  \includegraphics[width=\linewidth]{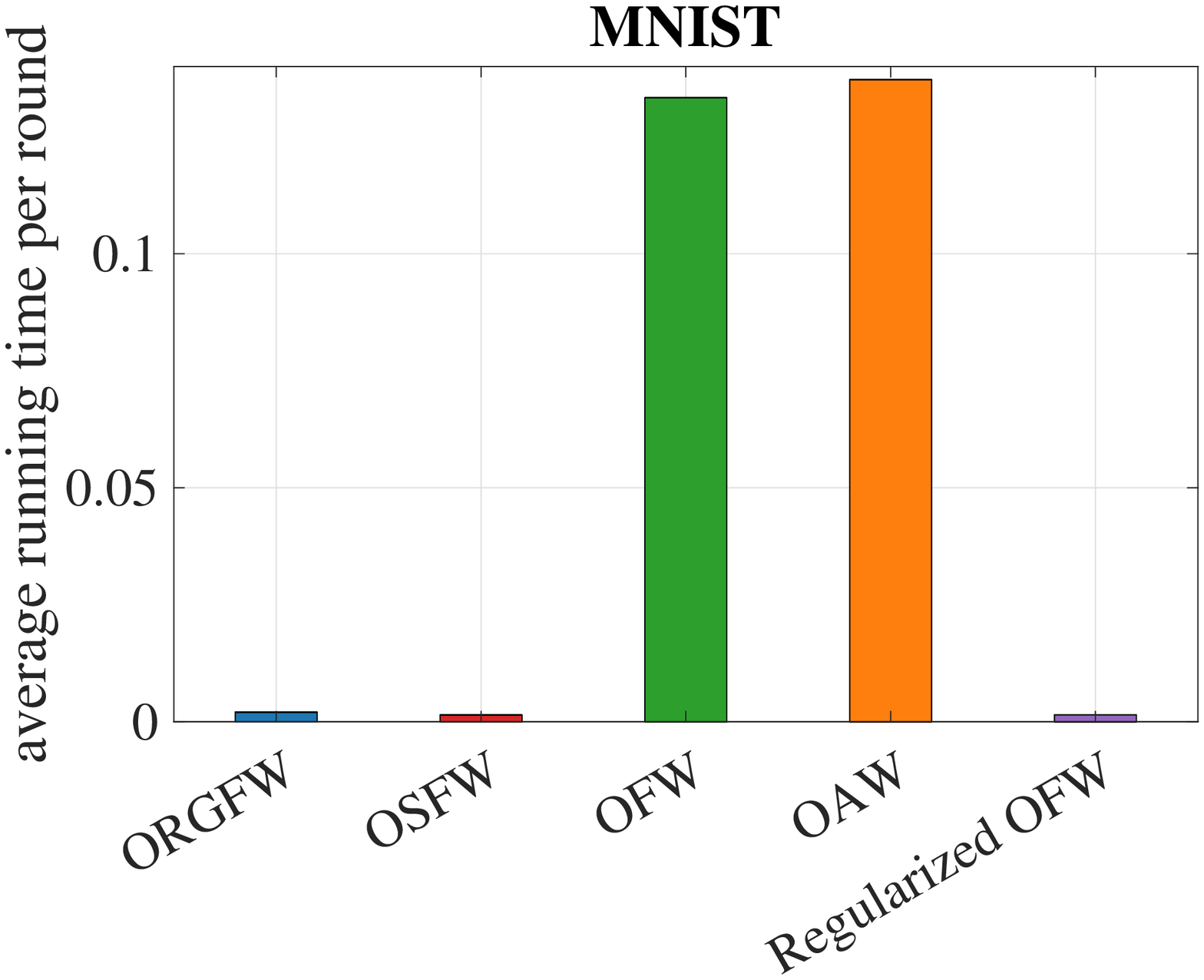}
\end{subfigure}
\hfill
\begin{subfigure}{.23\textwidth}
  \centering
  \includegraphics[width=\linewidth]{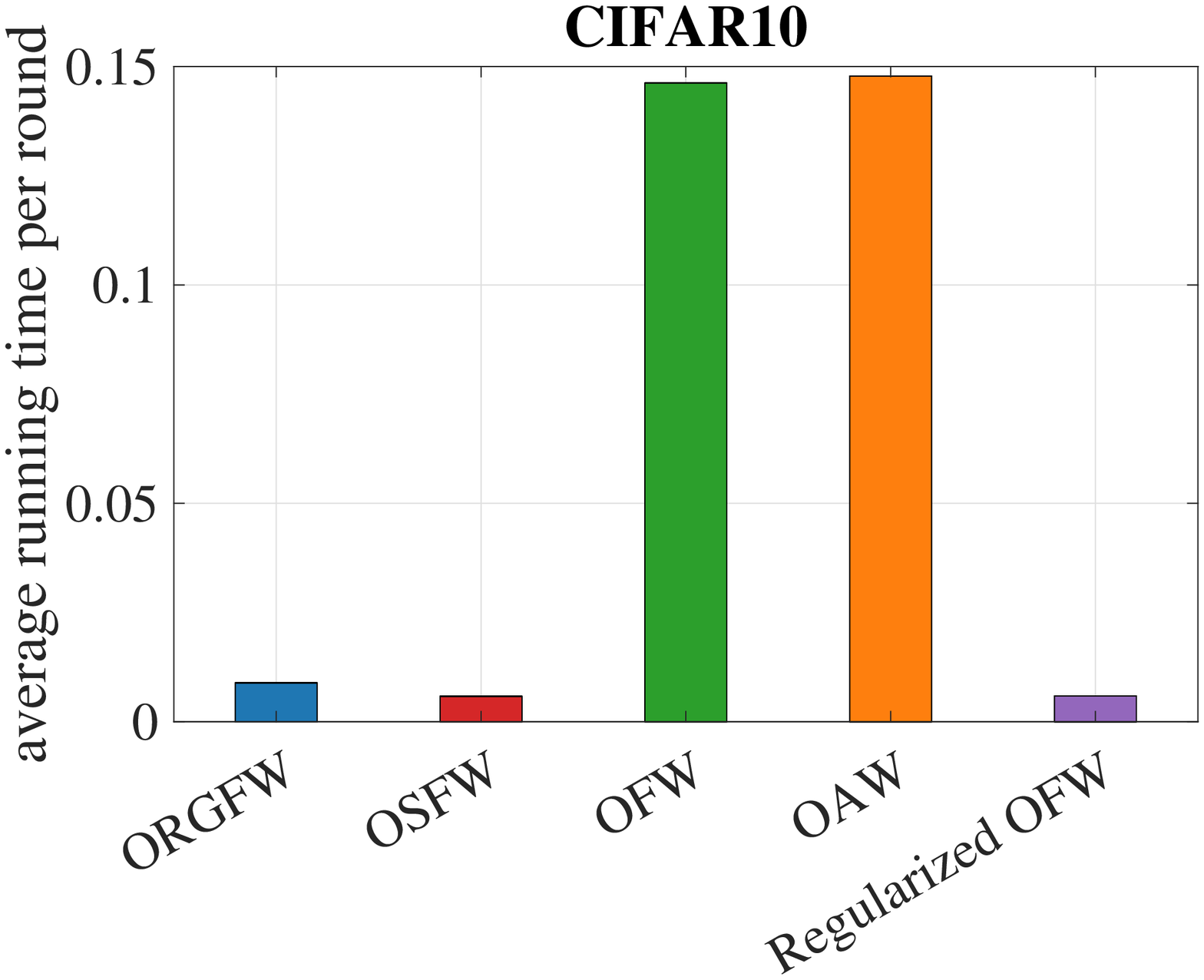}
\end{subfigure}
\\
\begin{subfigure}{.23\textwidth}
  \centering
  \includegraphics[width=\linewidth]{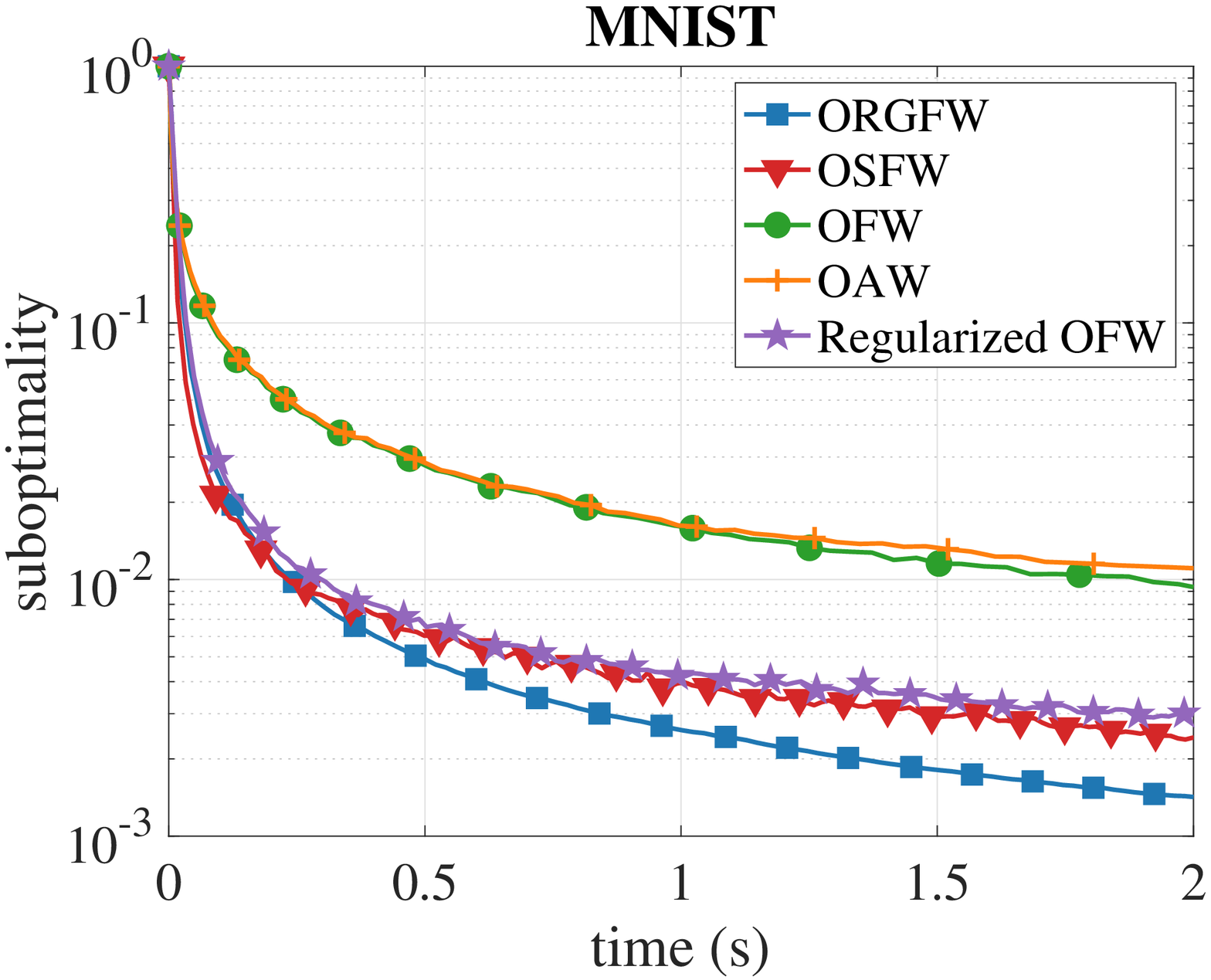}
\end{subfigure}
\hfill
\begin{subfigure}{.23\textwidth}
  \centering
  \includegraphics[width=\linewidth]{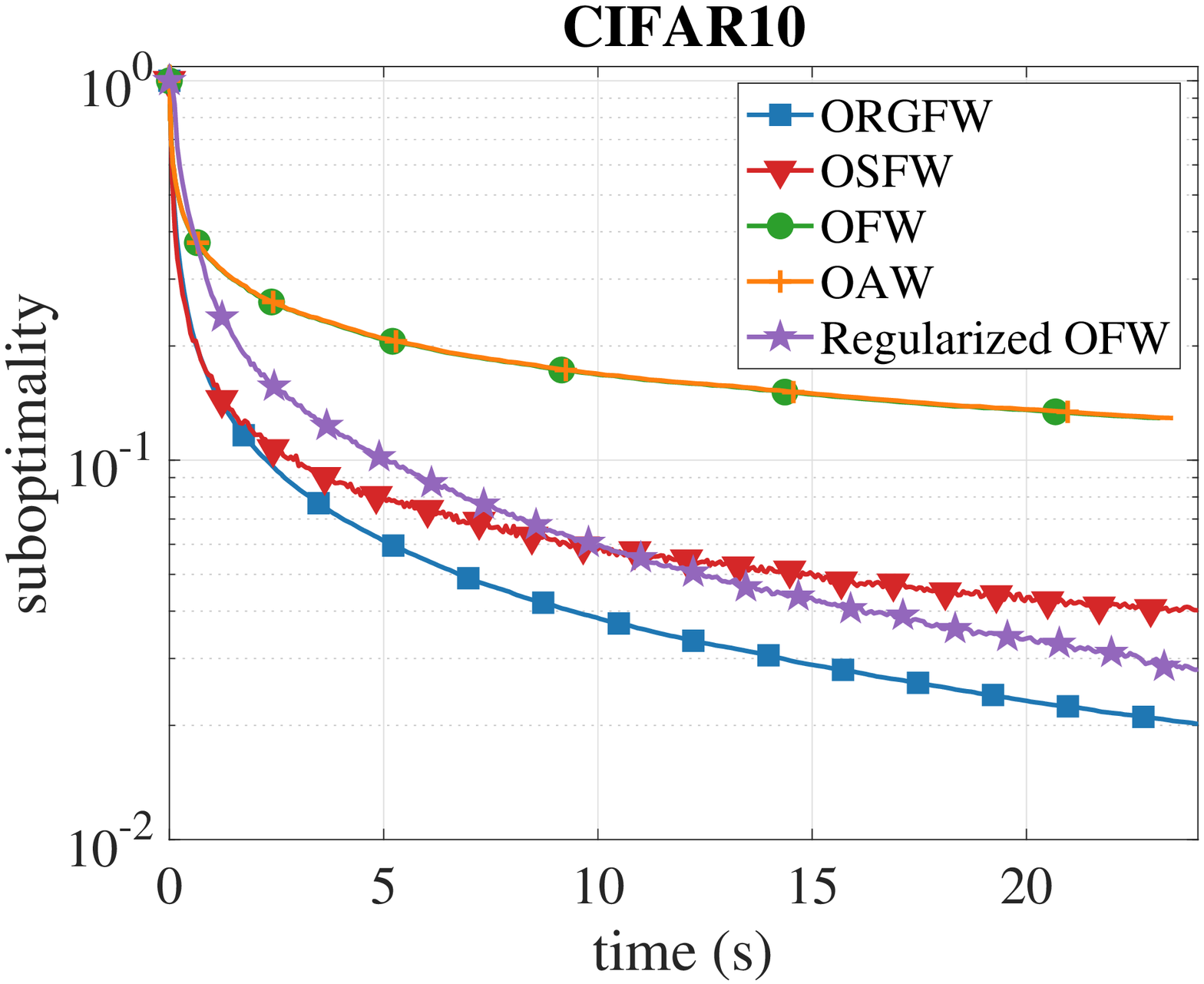}
\end{subfigure}
\caption{
Results on the online multiclass logistic regression task in the stochastic setting (left: MNIST, right: CIFAR10).
The left column shows the regret in the stochastic setting versus the number of rounds.
The right column shows the suboptimality $(\fbar(\WB_t) - \fbar(\WB^*)) / (\fbar(\WB_1) - \fbar(\WB^*))$, where $\WB^* \in \mathrm{argmin}_{\WB \in \CM} \fbar(\WB)$ and $\WB_1$ is the initial point.
}
\label{figure_LR_stochastic}
\end{figure}

\begin{figure}[t]
\begin{subfigure}{.23\textwidth}
  \centering
  \includegraphics[width=\linewidth]{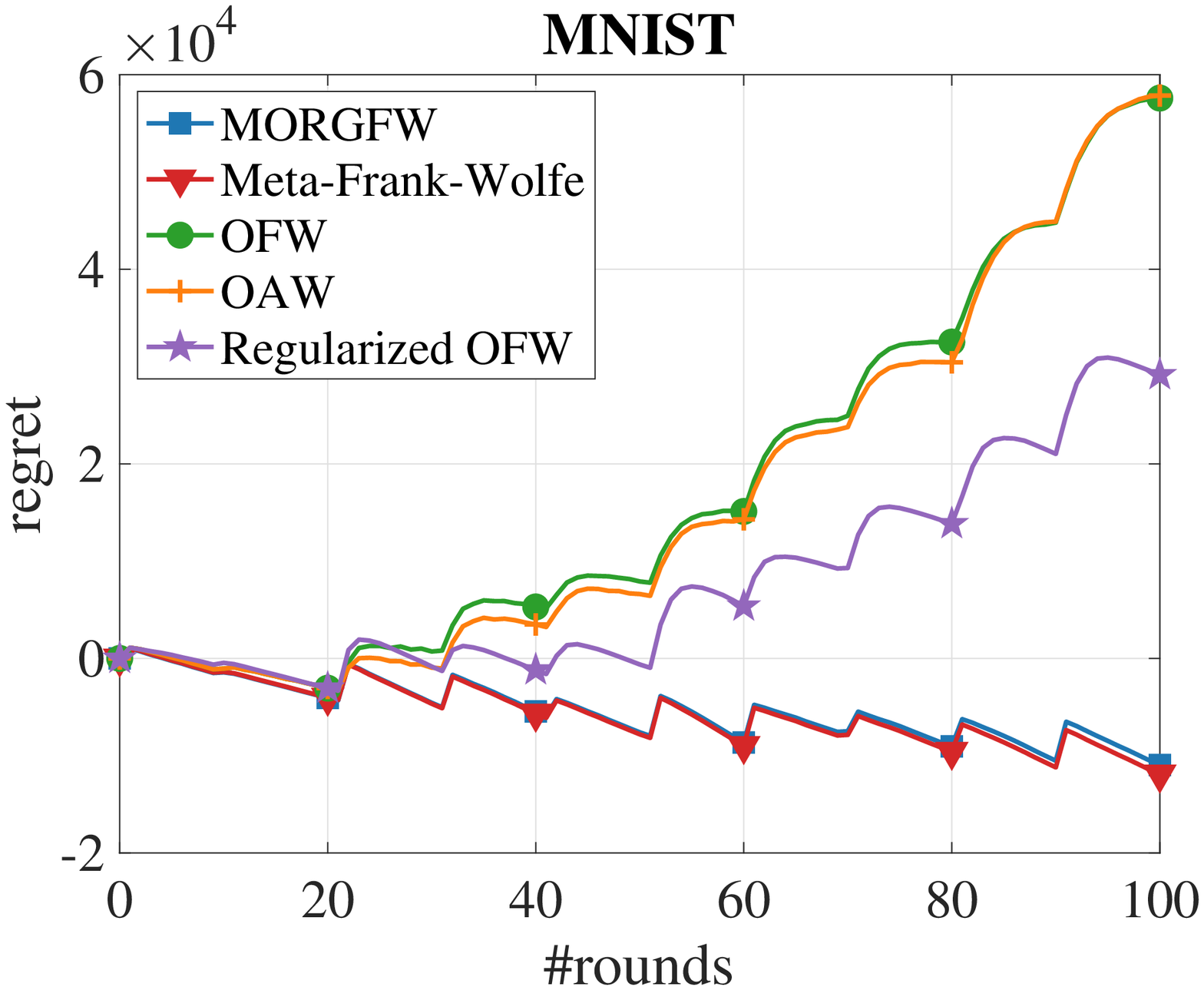}
\end{subfigure}
\hfill
\begin{subfigure}{.23\textwidth}
  \centering
  \includegraphics[width=\linewidth]{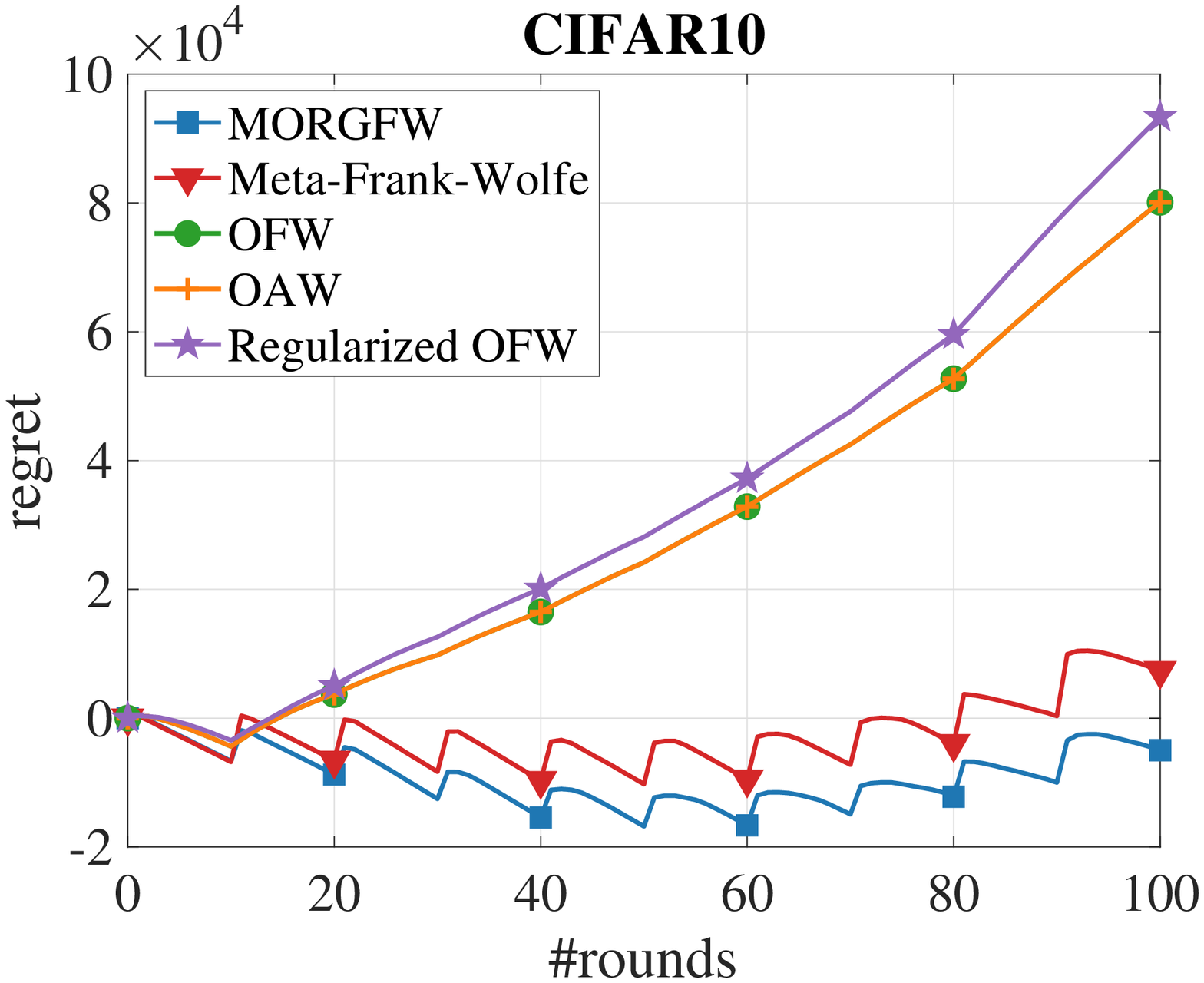}
\end{subfigure}
\\
\begin{subfigure}{.23\textwidth}
  \centering
  \includegraphics[width=\linewidth]{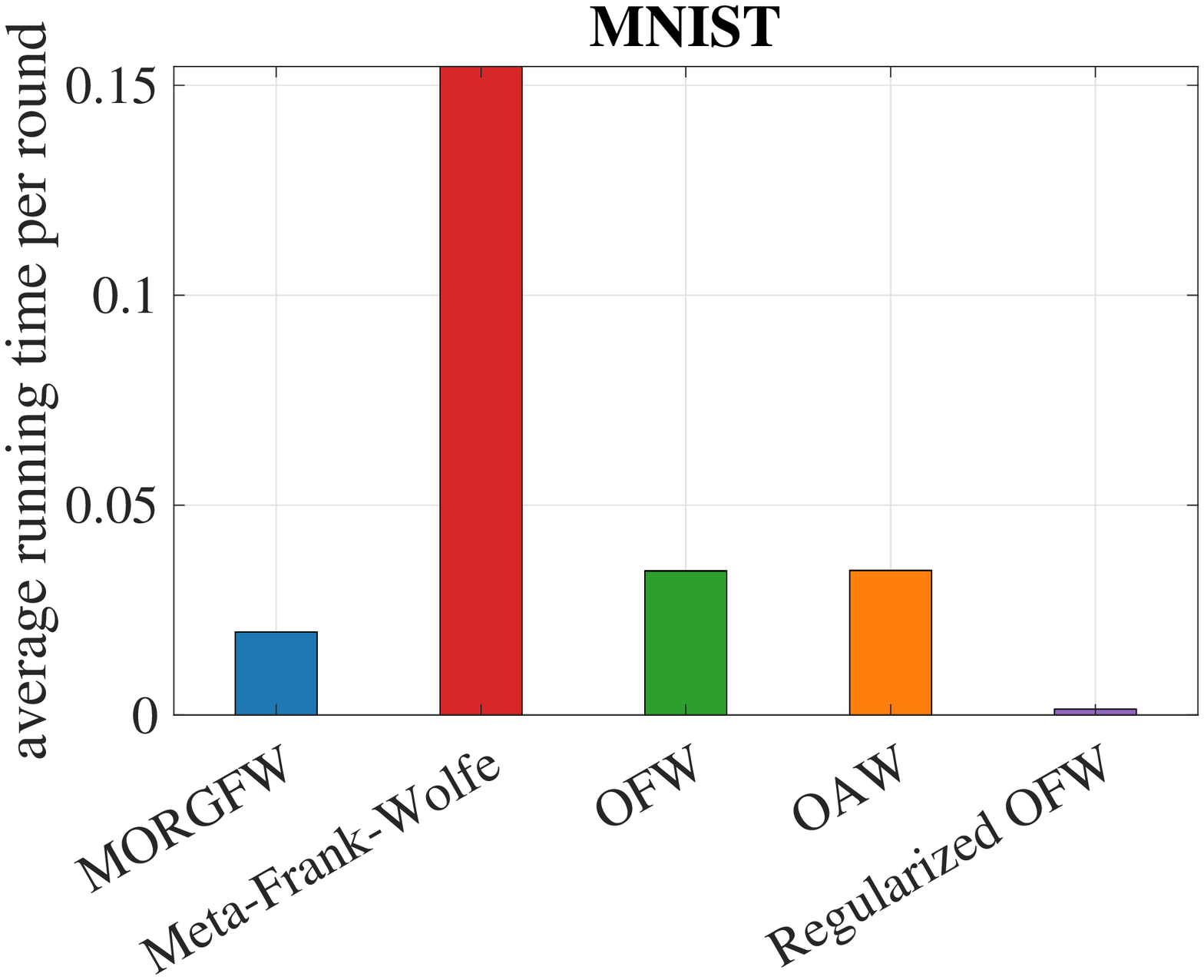}
\end{subfigure}
\hfill
\begin{subfigure}{.23\textwidth}
  \centering
  \includegraphics[width=\linewidth]{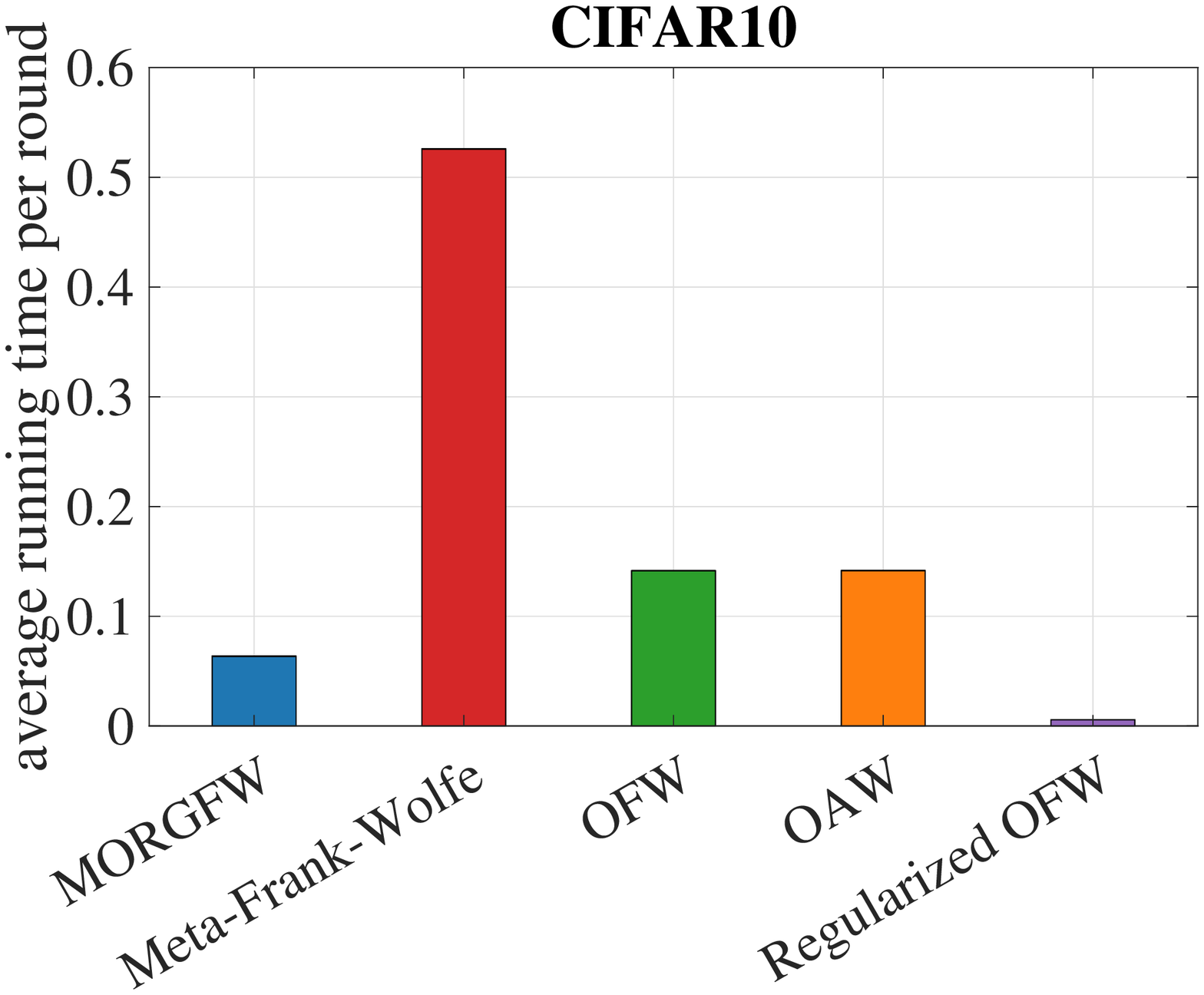}
\end{subfigure}
\caption{
Results on the online multiclass logistic regression task in the adversarial setting (left: MNIST, right: CIFAR10).
The left column shows the regret versus the number of rounds.
The right column shows the average running time per round of each method.
}
\label{figure_LR_adv}
\end{figure}

\begin{table}[t]
    \centering
	\caption{Summary of the multiclass datasets.}
	\label{table_datasets}
	\begin{tabular}{cccc}
		\toprule
		Dataset       & \#features  & \#instances & \#classes   \\
		\midrule
        MNIST         & $784$      & $60,000$   & $10$       \\
        CIFAR10       & $3072$     & $50,000$   & $10$       \\
		\bottomrule
	\end{tabular}
\end{table}

\begin{figure*}[ht]
\begin{subfigure}{.32\textwidth}
  \centering
  \includegraphics[width=\linewidth]{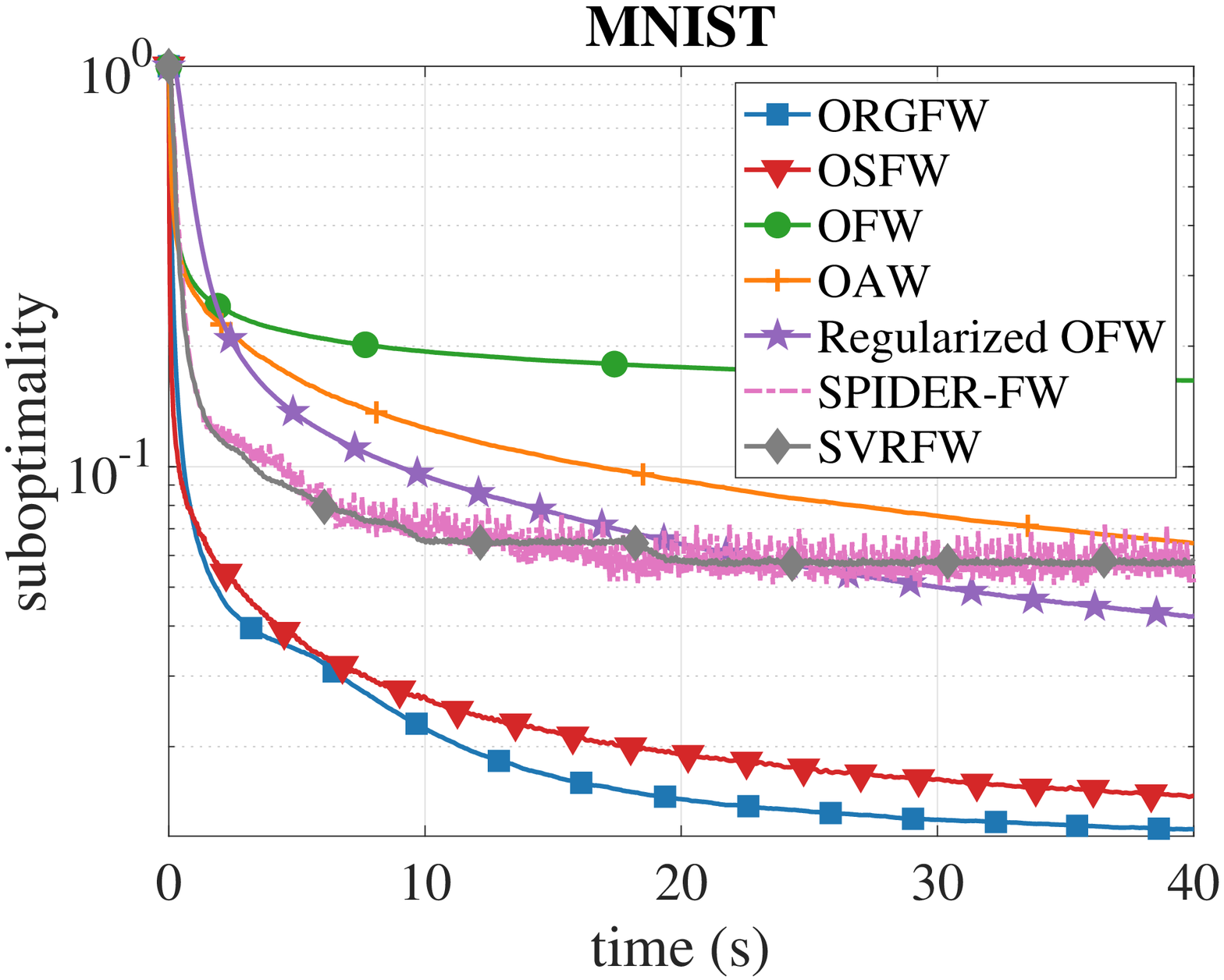}
\end{subfigure}
\hfill
\begin{subfigure}{.32\textwidth}
  \centering
  \includegraphics[width=\linewidth]{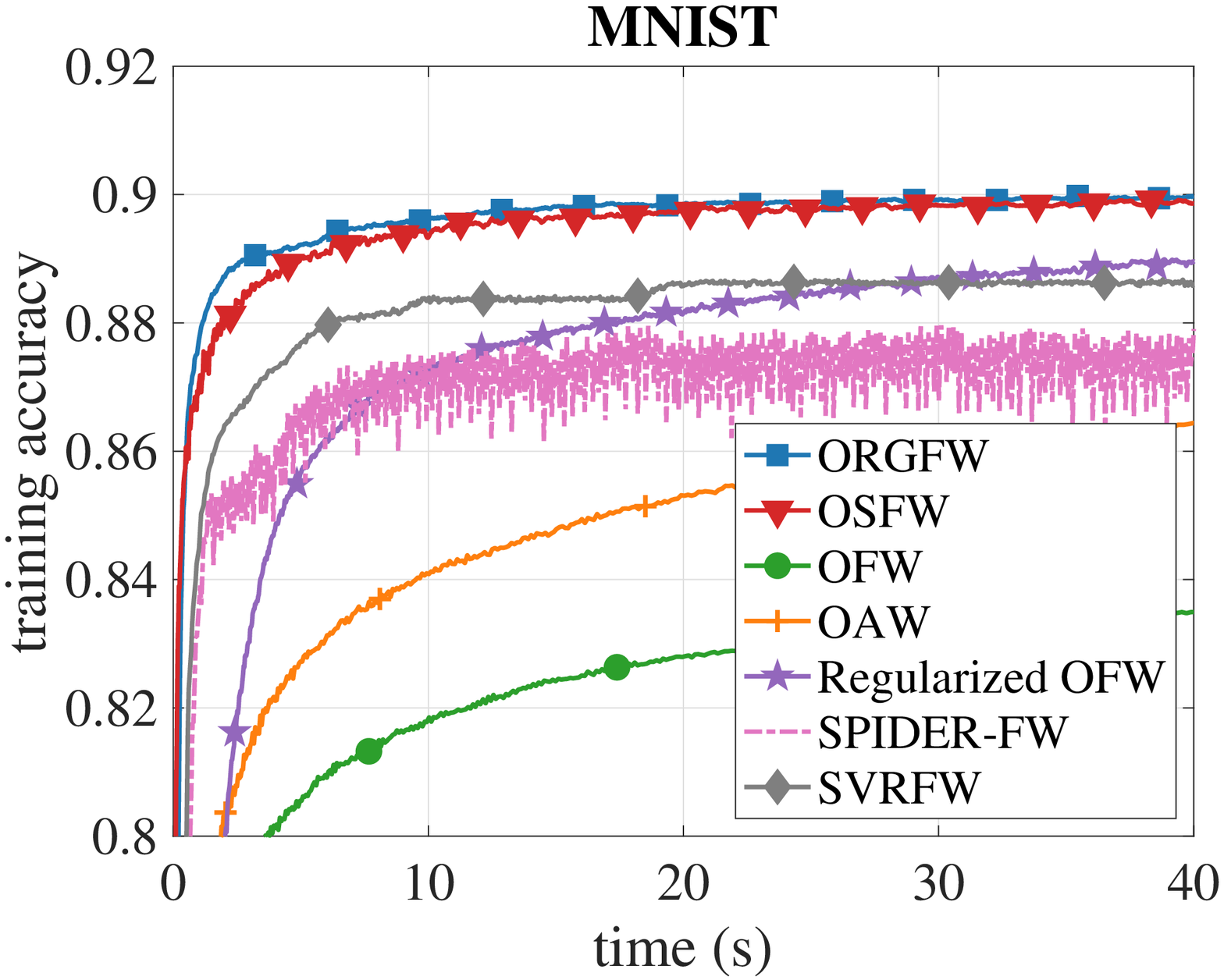}
\end{subfigure}
\hfill
\begin{subfigure}{.32\textwidth}
  \centering
  \includegraphics[width=\linewidth]{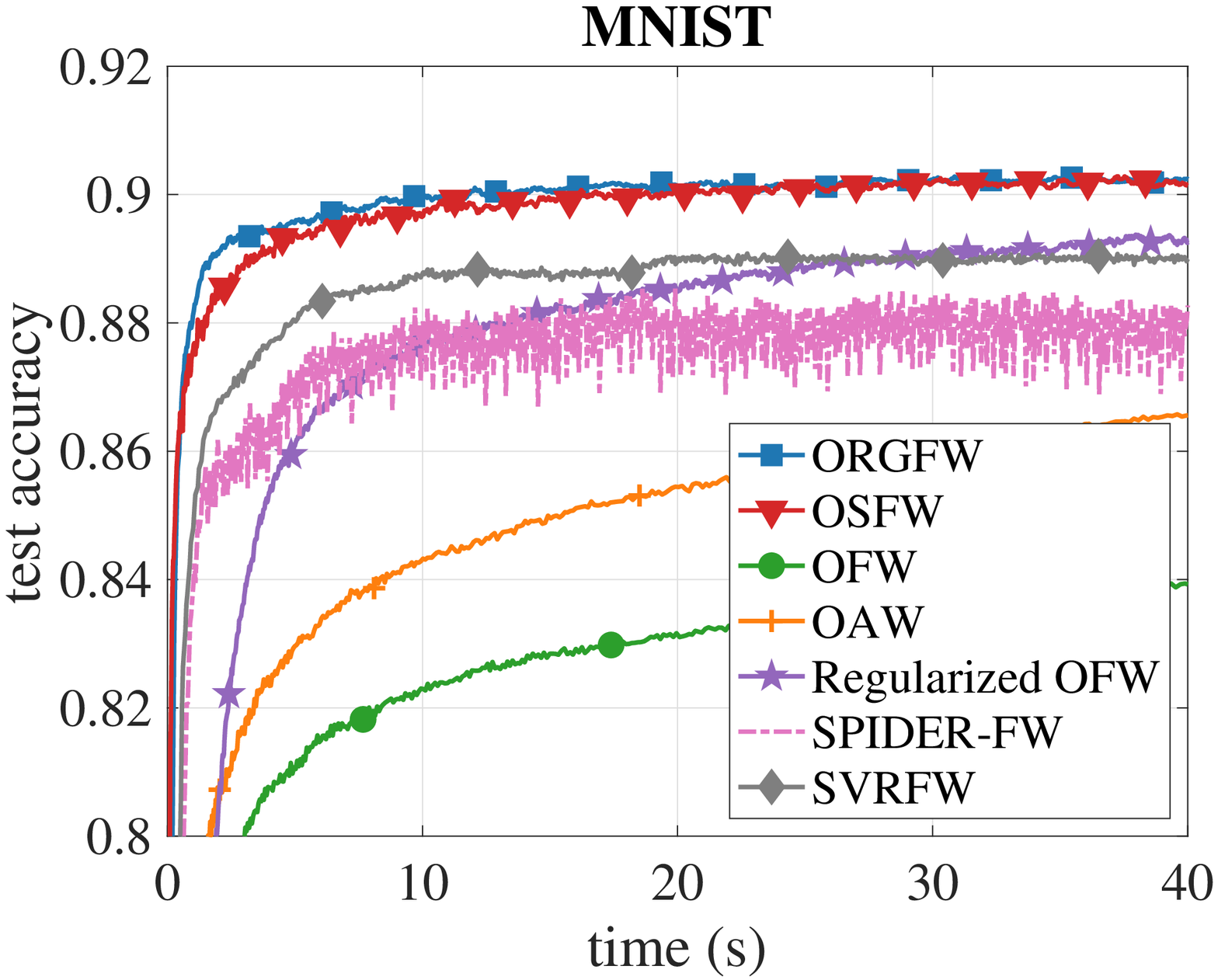}
\end{subfigure}
\\
\begin{subfigure}{.32\textwidth}
  \centering
  \includegraphics[width=\linewidth]{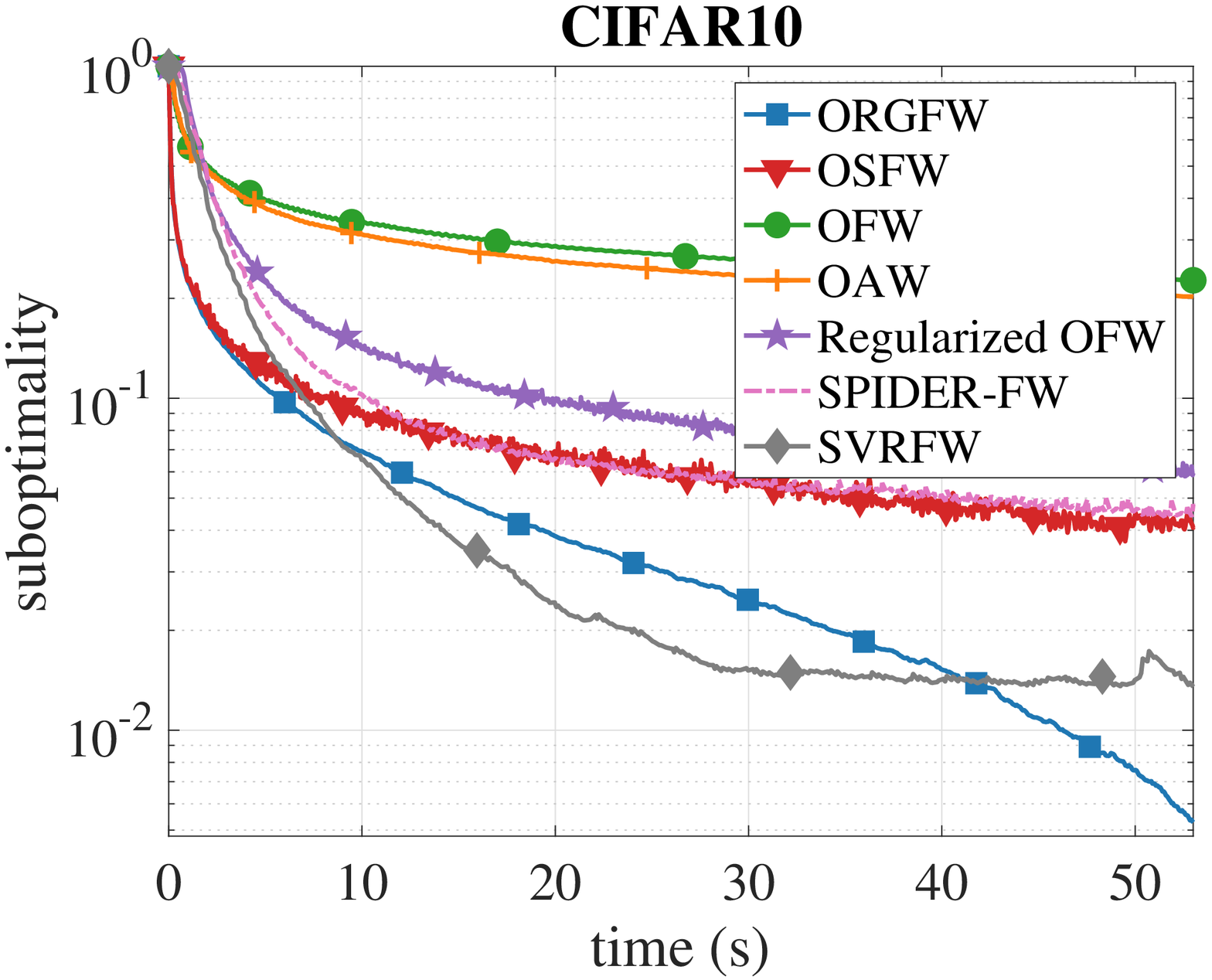}
\end{subfigure}
\hfill
\begin{subfigure}{.32\textwidth}
  \centering
  \includegraphics[width=\linewidth]{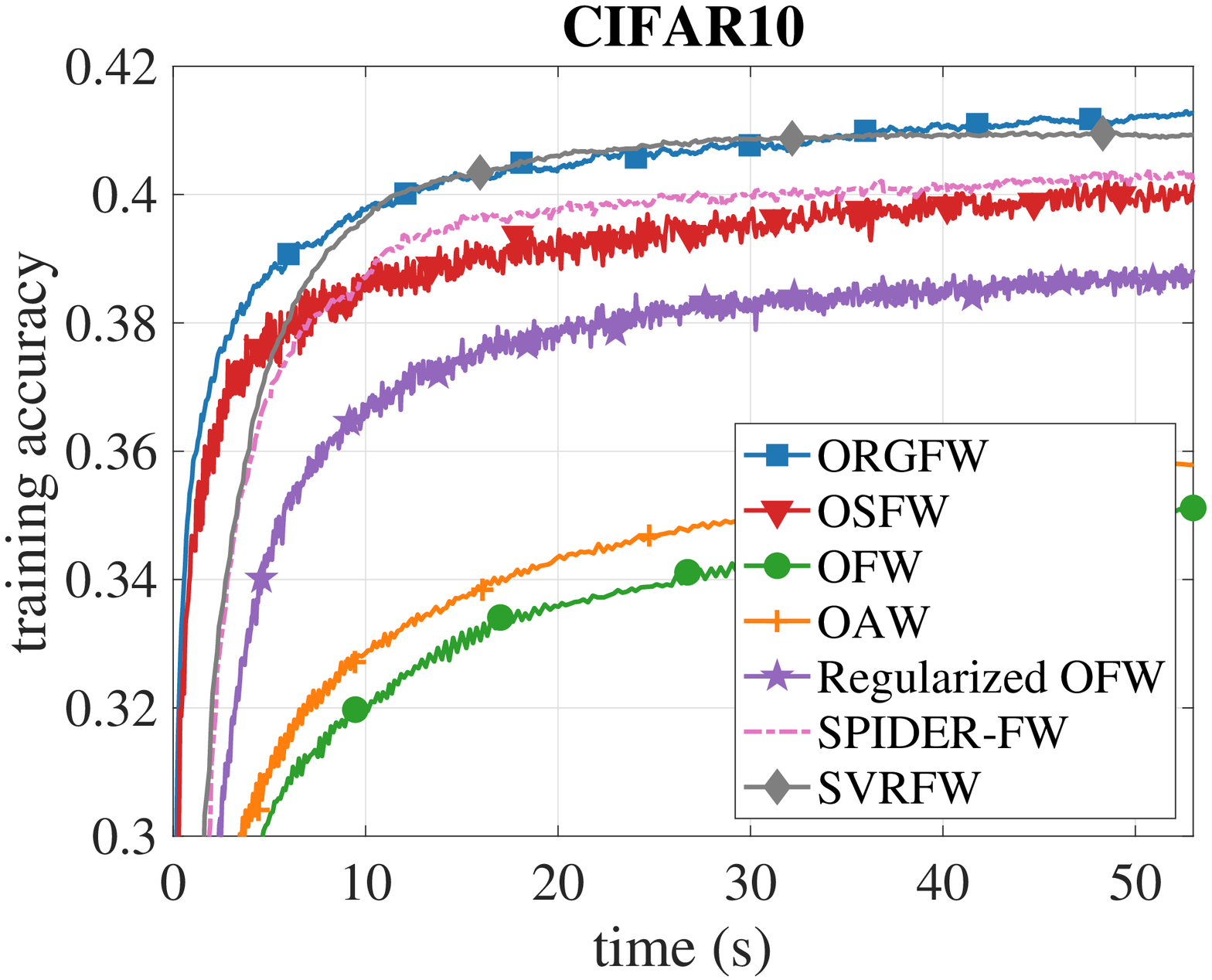}
\end{subfigure}
\hfill
\begin{subfigure}{.32\textwidth}
  \centering
  \includegraphics[width=\linewidth]{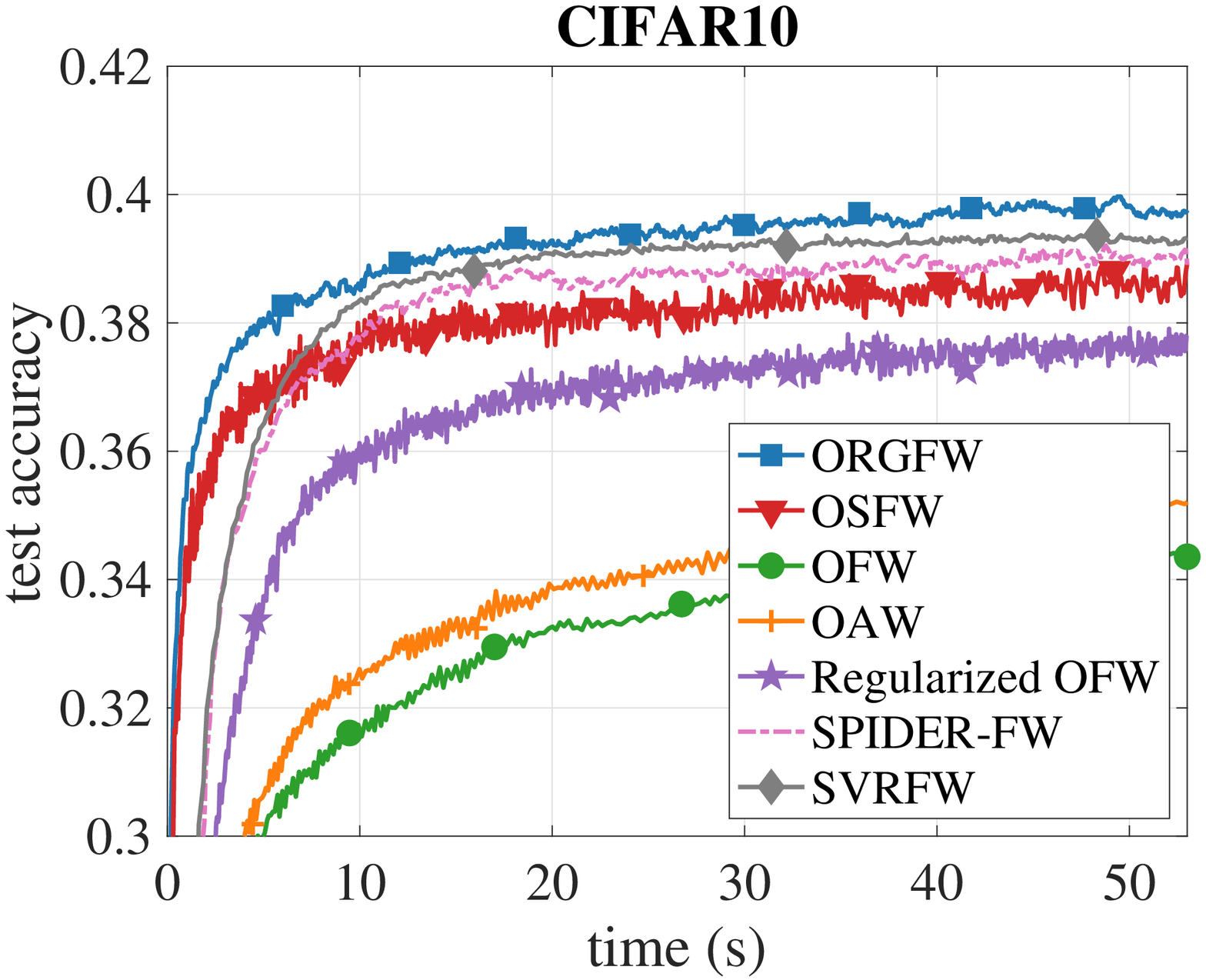}
\end{subfigure}
\caption{
Results on the one-hidden-layer neural network optimization problem (top: MNIST, bottom: CIFAR10).
The left column shows the suboptimality versus the running time.
The middle (resp., right) column shows the training (resp., test) accuracy versus the running time.
}
\label{figure_NN}
\end{figure*}

\subsection{Online Multiclass Logistic Regression}
In the first experiment, we consider an OCO problem -- online multiclass logistic regression~\cite{zhang2019quantized}.
In each round $t = 1, \ldots, T$, we receive a subset $\BM_t$ of data points with $|\BM_t| = B$, where each data point $i \in \BM_t$ is of the form $(\aB_i, y_i) \in \RBB^d \times \{1, \ldots, C\}$.
Here $\aB_i$ is a feature vector and $y_i \in \{1, \dots, C\}$ is the corresponding label.
We define $f_t$ as the multiclass logistic loss function
$$
\begin{aligned}
    f_t(\WB)
    = - \sum_{i \in \BM_t} \sum_{c=1}^C \oneB\{y_i = c\} \mathrm{log} \frac{\mathrm{exp}(\WB_c^T \aB_i)}{\sum_{j=1}^C \mathrm{exp}(\WB_j^T \aB_i)},
\end{aligned}
$$
and set the constraint $\CM = \{\WB \in \RBB^{d \times C}: \|\WB \|_1 \le r \}$ for some constant $r \in \RBB_+$, where $\|\WB\|_1$ denotes the matrix $\ell_1$ norm, i.e., $\|\WB\|_1 = \max_{1 \le j \le C} \sum_{i=1}^d |[\WB]_{ij}|$.
We note that the loss function $f_t$ is convex and smooth.
We consider both stochastic and adversarial online settings in this experiment.
In the stochastic setting, each subset of data $\BM_t$ is sampled i.i.d. from the whole dataset.
In the adversarial setting, we first sort data points by class label and then select $\{\BM_1, \ldots, \BM_T\}$ sequentially from these datasets after sorting.
For the MNIST dataset, we set $|\BM_t| = 600$ and $r = 8$.
For CIFAR10, we set $|\BM_t| = 500$ and $r = 32$.

In the stochastic setting, we compare the performance of ORGFW with OFW, Regularized OFW, OAW, and OSFW.
In the implementation of all these methods, we use the exact gradient $\nabla f_t$ in the $t$-th iteration since it can be computed efficiently.
We report the regret and the average per-iteration running time in the first two rows of Figure~\ref{figure_LR_stochastic}.
It can be seen from the top row that the regret of ORGFW is lower than OSFW and Regularized OFW, and slightly worse than OFW and OAW.
Nevertheless, the second row shows that the per-iteration computational cost of ORGFW is significantly better than OFW and OAW.
This implies that ORGFW has the advantages of low regret and low computational cost at the same time.
We also include the suboptimality in the third row of Figure~\ref{figure_LR_stochastic} to demonstrate the efficiency of ORGFW to solve the stochastic optimization problem $\min_{\WB \in \CM} \fbar(\WB) := \EBB_{\BM_t} [f_t(\WB)]$.
One can see that ORGFW outperforms all the other methods in terms of the suboptimality versus the running time.

In the adversarial setting, we compare MORGFW with OFW, Regularized OFW, OAW, and Meta-Frank-Wolfe.
We set the number of rounds to $T = 100$ and set the parameter $K$ in MORGFW and Meta-Frank-Wolfe to $K = T$ and $K = T^{3/2}$ as suggested by the theory, respectively.
The results are shown in Figure~\ref{figure_LR_adv}.
From the first row, we can see that the regret of MORGFW is comparable or lower than that of Meta-Frank-Wolfe and is significantly better than those of OFW, OAW, and Regularized OFW.
We note that the zig-zag phenomenon is due to
the adversarial nature of the loss function sequence.
From the second row, we can see that the per-round computational cost of MORGFW is only worse than Regularized OFW.
This confirms the advantages of MORGFW in achieving low regret and maintaining low computational cost simultaneously.

\subsection{Training a One-hidden-layer Neural Network}
In the second experiment, we focus on training a one-hidden-layer neural network with an additional $\ell_1$ norm constraint~\cite{zhang2019quantized}.
Specifically, given a multiclass data set $\{(\aB_i, y_i)\}_{i=1}^n$ with $(\aB_i, y_i) \in \RBB^d \times \{1, \ldots, C\}$,
we consider the following problem
$$
\begin{aligned}
    & \min_{\substack{\WB_1 \in \RBB^{d \times m}, \bB_1 \in \RBB^m \\ \WB_2 \in \RBB^{m \times C}, \bB_2 \in \RBB^C}}
    \EBB_{i}\Big[ h(y_i, \phi(\WB_2^T \sigma(\WB_1^T \aB_i + \bB_1) + \bB_2)) \Big],
\end{aligned}
$$
subject to $\| \WB_j \|_1 \le r_{w}, \ \| \bB_j \|_1 \le r_{b}, \ \forall j \in \{1, 2\}$.
Here, $i$ is a random variable sampled uniformly from $\{1, \dots, n\}$, $\phi$ is the softmax function, $\sigma(x) := 1/(1 + \mathrm{exp}(-x))$ is the sigmoid function, and $h(y, \pB) := - \sum_{c=1}^C \oneB \{y = c\} \mathrm{log} (\pB_c)$ for a probability vector $\pB = (\pB_1, \ldots, \pB_C)^T$.
We note that training a neural network subject to an $\ell_1$ constraint via FW-type methods exactly corresponds to a dropout regularization~\cite{ravi2019explicitly}.

We compare ORGFW with OSFW, Regularized OFW, OFW, and OAW.
We also include two state-of-the-art offline projection-free methods: SVRF~\cite{hazan2016variance,reddi2016stochastic} and SPIDER-FW~\cite{shen2019complexities}.
For the online methods, diminishing step sizes are used and
a mini-batch of $16$ data points are revealed to them in each round.
For SVRF and SPIDER-FW, we use constant step sizes as suggested by~\cite{shen2019complexities}.
We compare the performance of these methods on the MNIST and CIFAR10 datasets.
For both datasets, we set $m = 10$.
In addition, we set the $\ell_1$ ball radii $r_w = r_b = 10$.
The experimental results are shown in Figure~\ref{figure_NN}.
One can see that ORGFW has the best performance in terms of the suboptimality, the training accuracy, and the test accuracy.

\section{Conclusion}  \label{section_conclusion}

We proposed two efficient projection-free online methods, ORGFW and MORGFW, for solving online convex optimization problems in stochastic and adversarial settings, respectively.
We provided novel regret analysis, which shows that the proposed methods achieve nearly optimal $\tilde{\OM}(\sqrt{T})$ regret bounds with low computational costs.
In addition, we provided convergence analysis for ORGFW in stochastic optimization problems.
Experimental results validate the advantages of the proposed methods.

{
    \small
    \bibliography{references}
    \bibliographystyle{aaai}
}

\onecolumn
\appendix
\section{Deferred Proofs} \label{section_proofs}

In this section, we provide detailed proofs of lemmas and theorems in Section~\ref{section_analysis}.
For a sequence of real numbers $\{\alpha_t\}$,
we use the convention that $\prod_{k = \tau}^{t} \alpha_k = 1$ and $\sum_{k = \tau}^{t} \alpha_k = 0$ if $\tau > t$.
We first present a useful Azuma-Hoeffding-type concentration inequality for vector valued martingales~\cite{pinelis1994optimum}.
See also~\cite{fang2018spider} and references therein.
\begin{proposition}  \label{proposition_concentration}
\cite[Theorem 3.5]{pinelis1994optimum}
Let $\zetaB_1, \zetaB_2, \ldots, \zetaB_t \in \RBB^d$ be a vector-valued martingale difference sequence w.r.t. a filtration $\{\FM_{t}\}$, i.e., for each $\tau \in 1, \ldots, t$, we have $\EBB[ \zetaB_{\tau} | \FM_{\tau - 1}] = \zeroB$.
Suppose that $\| \zetaB_{\tau} \| \le c_{\tau}$ almost surely.
Then, $\forall t \ge 1$,
\begin{equation}
    \PBB \Big( \Big\| \sum_{\tau = 1}^t \zetaB_{\tau} \Big\| \ge \lambda \Big)
    \le 4 \mathrm{exp}(- \frac{\lambda^2}{4 \sum_{\tau = 1}^t c_{\tau}^2}).
\end{equation}
\end{proposition}

\subsection{Proof of Lemma~\ref{lemma_gradient_error_bound_whp}}  \label{section_proof_key_lemma}
Before we proceed to the proof of Lemma~\ref{lemma_gradient_error_bound_whp}, we present and prove the following technical lemma, which characterizes the convergence behavior of a sequence $s_t$.
\begin{lemma}  \label{lemma_sequence_convergence}
    Define $\rho_k = 1/(k+1)^{\alpha}$ where $\alpha \in (0, 1]$ and $k \ge 1$.
    Let $\{s_t\}$ be a sequence of real numbers satisfying
    \begin{equation}
        s_t = \sum_{\tau = 2}^t \Big( \rho_{\tau - 1} \prod_{k = \tau}^t (1 - \rho_k) \Big)^2,
    \end{equation}
    for all $t \ge 2$.
    Then the sequence $\{ s_t \}$ converges to zero at the rate
    \begin{equation}  \label{eq_sequence_convergence}
        s_t \le \frac{1}{(t + 1)^{\alpha}}.
    \end{equation}
\end{lemma}

\begin{proof}
    We prove the lemma by induction.
    For $t = 2$, we observe that
    \begin{equation}
        s_2
        = (\frac{1}{2^{\alpha}} \cdot \frac{3^{\alpha} - 1}{3^{\alpha}})^2
        \le (\frac{1}{2^{\alpha}} \cdot \frac{2^{\alpha}}{3^{\alpha}})^2
        = \frac{1}{9^{\alpha}} \le \frac{1}{3^{\alpha}},
    \end{equation}
    where the first inequality follows from the concavity of the function  $h(x) = x^{\alpha}$, i.e., $(x + 1)^{\alpha} \le x^{\alpha} + 1$ for any $x \ge 0$ and $\alpha \in (0, 1]$.
    Now we suppose that~\eqref{eq_sequence_convergence} holds when $t = T$ for some $T \ge 2$, i.e.,
    \begin{equation}  \label{eq_sequence_convergence_hypothesis}
        s_T = \sum_{\tau = 2}^T \Big( \rho_{\tau - 1} \prod_{k = \tau}^T (1 - \rho_k) \Big)^2 \le \frac{1}{(T + 1)^{\alpha}}.
    \end{equation}
    For $t = T + 1$, we have
    \begin{equation}
    \begin{aligned}[b]
        s_{T + 1}
        &= \sum_{\tau = 2}^{T + 1} \Big( \rho_{\tau - 1} \prod_{k = \tau}^{T + 1} (1 - \rho_k) \Big)^2
        = \sum_{\tau = 2}^{T + 1} \Big( \rho_{\tau - 1} (1 - \rho_{T + 1}) \prod_{k = \tau}^{T} (1 - \rho_k) \Big)^2  \\
        &= (1 - \rho_{T + 1})^2 \Big( \sum_{\tau = 2}^{T} \big( \rho_{\tau - 1} \prod_{k = \tau}^{T} (1 - \rho_k) \big)^2  + \rho_{T}^2 \Big)
        = (1 - \rho_{T + 1})^2 ( s_T  + \rho_{T}^2 )  \\
        &\overset{(a)}{\le} \Big( \frac{(T + 2)^{\alpha} - 1}{(T + 2)^{\alpha}} \Big)^2 \Big( \frac{1}{(T + 1)^{\alpha}}  +  \frac{1}{(T + 1)^{2 \alpha}} \Big)
        = \frac{((T + 2)^{\alpha} - 1)^2 ((T + 1)^{\alpha} + 1)}{(T + 2)^{2 \alpha} (T + 1)^{2 \alpha}}  \\
        &\overset{(b)}{\le} \frac{((T + 2)^{\alpha} - 1) ((T + 1)^{\alpha} + 1)}{(T + 2)^{2 \alpha} (T + 1)^{\alpha}}
        = \frac{(T + 2)^{\alpha}(T + 1)^{\alpha} + (T + 2)^{\alpha} - 1 - (T + 1)^{\alpha}}{(T + 2)^{2 \alpha} (T + 1)^{\alpha}}  \\
        &\overset{(c)}{\le} \frac{(T + 2)^{\alpha}(T + 1)^{\alpha}}{(T + 2)^{2 \alpha} (T + 1)^{\alpha}}
        = \frac{1}{(T + 2)^{\alpha}}.
    \end{aligned}
    \end{equation}
    where (a) follows from the induction hypothesis~\eqref{eq_sequence_convergence_hypothesis} and the definition of $\rho_k$;
    (b) and (c) follow from the concavity of the scalar function $h(x) = x^{\alpha}$.
    This completes the induction step.
    Therefore, we have $s_t \le 1/(t + 1)^{\alpha}$ for any $t \ge 2$.
\end{proof}

Having established the above lemma, we proceed to prove Lemma~\ref{lemma_gradient_error_bound_whp}.
\begin{proof}
(Proof of Lemma~\ref{lemma_gradient_error_bound_whp})
We first reformulate $\epsilonB_t$ as the sum of a martingale difference sequence.
For $t > 1$, we have
\begin{equation}
\begin{aligned}[b]
    \epsilonB_t
    &= (1 - \rho_t) \epsilonB_{t-1}
        + \rho_t \big( \nabla F_t(\xB_t, \xi_t) - \nabla \fbar(\xB_t) \big) \\
        &\ \ \ \ + (1 - \rho_t) \Big( \nabla F_t(\xB_t, \xi_t) - \nabla F_t(\xB_{t-1}, \xi_t) - \big( \nabla \fbar(\xB_t) - \nabla \fbar(\xB_{t-1}) \big) \Big)   \\
    &= \prod_{k = 2}^t (1 - \rho_{k}) \epsilonB_{1}
        + \sum_{\tau = 2}^t \prod_{k = \tau}^t (1 - \rho_{k}) \Big( \nabla F_{\tau}(\xB_{\tau}, \xi_{\tau}) - \nabla F_{\tau}(\xB_{\tau-1}, \xi_{\tau}) - \big( \nabla \fbar(\xB_{\tau}) - \nabla \fbar(\xB_{\tau-1}) \big) \Big)   \\
        &\ \ \ \ + \sum_{\tau = 2}^t \rho_{\tau} \prod_{k = \tau + 1}^{t} (1 - \rho_k) \big( \nabla F_{\tau}(\xB_{\tau}, \xi_{\tau}) - \nabla \fbar(\xB_{\tau}) \big).
\end{aligned}
\end{equation}
We let $\epsilonB_t = \sum_{\tau = 1}^t \zetaB_{t, \tau}$, where
$\zetaB_{t, 1} = \prod_{k = 2}^t (1 - \rho_{k}) \epsilonB_{1}$ and
$\zetaB_{t, \tau} =
    \prod_{k = \tau}^t (1 - \rho_{k}) \big( \nabla F_{\tau}(\xB_{\tau}, \xi_{\tau}) - \nabla F_{\tau}(\xB_{\tau-1}, \xi_{\tau}) - \big( \nabla \fbar(\xB_{\tau}) - \nabla \fbar(\xB_{\tau-1}) \big) \big)
    + \rho_{\tau} \prod_{k = \tau + 1}^{t} (1 - \rho_k) \big( \nabla F_{\tau}(\xB_{\tau}, \xi_{\tau}) - \nabla \fbar(\xB_{\tau}) \big)$ for $\tau > 1$.
Recall that $\epsilonB_1 = \nabla F_1(\xB_1, \xi_1) - \nabla \fbar(\xB_1)$.
We observe that $\EBB[\zetaB_{t, \tau} | \FM_{\tau - 1}] = \zeroB$ where $\FM_{\tau - 1}$ is the $\sigma$-field generated by $\{f_1, \xi_1, \ldots, f_{\tau - 1}, \xi_{\tau - 1}\}$.
Therefore, $\{\zetaB_{t, \tau}\}_{\tau = 1}^t$ is a martingale difference sequence.

In what follows, we derive upper bounds of $\|\zetaB_{t, \tau}\|$.
We start by observing that for any $\tau = 1, 2, \dots, t$,
\begin{equation}
    \prod_{k = \tau}^{t} (1 - \rho_k)
    = \prod_{k = \tau}^{t} (1 - \frac{1}{(k+1)^{\alpha}})
    = \prod_{k = \tau}^{t} \frac{(k + 1)^{\alpha} - 1}{(k + 1)^{\alpha}}
    \le \prod_{k = \tau}^{t} \frac{k^{\alpha}}{(k + 1)^{\alpha}}
    = \frac{\tau^{\alpha}}{(t + 1)^{\alpha}},
\end{equation}
where the inequality follows from the concavity of $h(x) = x^{\alpha}$ for any $x \ge 0$.
By using the above inequality, we can bound $\|\zetaB_{t, 1}\|$ as follows
\begin{equation}  \label{eq_zeta_1_bound}
    \| \zetaB_{t, 1} \|
    \le \frac{2^{\alpha}}{(t+1)^{\alpha}} \| \nabla F_1(\xB_1, \xi_1) - \nabla \fbar(\xB_1) \|
    \le \frac{2^{\alpha} \sigma}{(t + 1)^{\alpha}}
    \defi c_{t, 1},
\end{equation}
where the second inequality follows from Assumption~\ref{assumption_stoch}.\ref{assumption_stoch_gradient}.
For $\tau > 1$, $\| \zetaB_{t, \tau} \|$ can be bounded by
\begin{equation}  \label{eq_zeta_tau_bound_pre}
\begin{aligned}[b]
    \| \zetaB_{t, \tau} \|
    &\le \prod_{k = \tau}^{t} (1 - \rho_k) \big( \| \nabla F_{\tau}(\xB_{\tau}, \xi_{\tau}) - \nabla F_{\tau}(\xB_{\tau - 1}, \xi_{\tau}) \| + \| \nabla \fbar(\xB_{\tau}) - \nabla \fbar(\xB_{\tau - 1}) \| \big) \\
    & \ \ \ \ + \rho_{\tau} \prod_{k = \tau + 1}^{t} (1 - \rho_k)
    \| \nabla F_{\tau}(\xB_{\tau}, \xi_{\tau}) - \nabla \fbar(\xB_{\tau}) \| \\
    &\overset{(a)}{\le} 2 L \| \xB_{\tau} - \xB_{\tau - 1} \| \prod_{k = \tau}^{t} (1 - \rho_k)
        + \sigma \rho_{\tau} \prod_{k = \tau + 1}^{t} (1 - \rho_k) \\
    &= 2 L \eta_{\tau - 1} \| \vB_{\tau - 1} - \xB_{\tau - 1} \| \prod_{k = \tau}^{t} (1 - \rho_k)
        + \sigma \rho_{\tau} \prod_{k = \tau + 1}^{t} (1 - \rho_k) \\
    &\overset{(b)}{\le} 2 L D \rho_{\tau - 1} \prod_{k = \tau}^{t} (1 - \rho_k)
        + \sigma \rho_{\tau} \prod_{k = \tau + 1}^{t} (1 - \rho_k).
\end{aligned}
\end{equation}
where (a) follows from Assumption~\ref{assumption_stoch};
(b) follows from the condition $\eta_k = \rho_k$ and Assumption~\ref{assumption_constraint}.
We observe that
\begin{equation}  \label{eq_zeta_tau_common_coefficient}
\begin{aligned}[b]
    \rho_{\tau} \prod_{k = \tau + 1}^{t} (1 - \rho_k)
    &= \frac{\rho_{\tau}}{\rho_{\tau - 1} (1 - \rho_{\tau})} \big( \rho_{\tau - 1} \prod_{k = \tau}^{t} (1 - \rho_k) \big)
    \le \frac{1}{1 - \rho_{\tau}} \big( \rho_{\tau - 1} \prod_{k = \tau}^{t} (1 - \rho_k) \big) \\
    &\le \frac{1}{1 - 1/3^{\alpha}} \big( \rho_{\tau - 1} \prod_{k = \tau}^{t} (1 - \rho_k) \big)
    \le \frac{3^{\alpha}}{3^{\alpha} - 1} \big( \rho_{\tau - 1} \prod_{k = \tau}^{t} (1 - \rho_k) \big).
\end{aligned}
\end{equation}
Plugging~\eqref{eq_zeta_tau_common_coefficient} into~\eqref{eq_zeta_tau_bound_pre}, we have, $\forall \tau > 1$,
\begin{equation}  \label{eq_zeta_tau_bound}
    \| \zetaB_{t, \tau} \|
    \le (2 L D + \frac{3^{\alpha} \sigma}{3^{\alpha} - 1}) \rho_{\tau - 1} \prod_{k = \tau}^{t} (1 - \rho_k)
    \defi c_{t, \tau}.
\end{equation}
Hence, by Proposition~\ref{proposition_concentration}, we have for any $\lambda \ge 0$,
\begin{equation}  \label{eq_concentration}
    \PBB \Big( \| \epsilonB_t \| \ge \lambda \Big)
    \le 4 \mathrm{exp} \Big( - \frac{\lambda^2}{4\sum_{\tau = 1}^t c_{t, \tau}^2} \Big),
\end{equation}
where $c_{t, 1}$ is defined in~\eqref{eq_zeta_1_bound} and $c_{t, \tau}$ for $\tau > 1$ is defined in~\eqref{eq_zeta_tau_bound}.
We can bound $\sum_{\tau = 1}^t c_{t, \tau}^2$ using Lemma~\ref{lemma_sequence_convergence} as follows
\begin{equation}  \label{eq_sum_c_sqr}
\begin{aligned}[b]
    \sum_{\tau = 1}^t c_{t, \tau}^2
    &= c_{t, 1}^2 + \sum_{\tau = 2}^t c_{t, \tau}^2
    = \frac{2^{2 \alpha} \sigma^2}{(t + 1)^{2 \alpha}} + (2 L D + \frac{3^{\alpha} \sigma}{3^{\alpha} - 1})^2 \sum_{\tau = 2}^t \big( \rho_{\tau - 1} \prod_{k = \tau}^{t} (1 - \rho_k) \big)^2  \\
    &\le \frac{2^{2 \alpha} \sigma^2}{(t + 1)^{2 \alpha}} + \frac{(2 L D + \frac{3^{\alpha} \sigma}{3^{\alpha} - 1})^2 }{(t + 1)^{\alpha}}
    \le \frac{((\sqrt{2})^{\alpha} \sigma)^2}{(t + 1)^{\alpha}} + \frac{(2 L D + \frac{3^{\alpha} \sigma}{3^{\alpha} - 1})^2 }{(t + 1)^{\alpha}}
    \le \frac{2 (2 L D + \frac{3^{\alpha} \sigma}{3^{\alpha} - 1})^2 }{(t + 1)^{\alpha}},
\end{aligned}
\end{equation}
where the last inequality follows from the fact that $(\sqrt{2})^{\alpha} \le 3^{\alpha} / (3^{\alpha} -1)$ for any $\alpha \in (0, 1]$.
Plugging~\eqref{eq_sum_c_sqr} into~\eqref{eq_concentration} and setting $\lambda = 2 (2 L D + \frac{3^{\alpha} \sigma}{3^{\alpha} - 1})(t + 1)^{- \alpha / 2} \sqrt{2 \mathrm{log}(4 / \delta_0)}$ for some $\delta_0 \in (0, 1)$, we have with probability at least $1 - \delta_0$,
\begin{equation}
    \| \epsilonB_t \| \le 2 (2 L D + \frac{3^{\alpha} \sigma}{3^{\alpha} - 1})(t + 1)^{- \alpha / 2} \sqrt{2 \mathrm{log}(4 / \delta_0)},
\end{equation}
which is the desired result.
\end{proof}

\subsection{Proof of Theorem~\ref{theorem_convex}}
To begin with, we present a useful lemma.
\begin{lemma} \label{lemma_descent_convex}
    \cite[Lemma 2]{mokhtari2018stochastic}
    Consider the proposed ORGFW method (Algorithm~\ref{algorithm_ORGFW}).
    If the expected objective function $\fbar$ is convex and the conditions in Assumptions~\ref{assumption_constraint} and~\ref{assumption_stoch_smooth} are satisfied, then
    \begin{equation}
        \fbar(\xB_{t+1}) - \fbar(\xB^*) \le (1 - \eta_t) \big( \fbar(\xB_{t}) - \fbar(\xB^*) \big) + \eta_t D \| \epsilonB_t \| + \frac{L D^2 \eta_t^2}{2}.
    \end{equation}
\end{lemma}

Lemma~\ref{lemma_descent_convex} basically shows that the loss function value $\fbar(\xB_{t})$ converges to $\fbar(\xB^*)$ as long as the error $\| \epsilonB_t \|$ can be properly controlled.
Now we are ready to prove Theorem~\ref{theorem_convex}.
\begin{proof}
(Proof of Theorem~\ref{theorem_convex})
We first construct a sequence $s_t = f_t(\xB_t) - f_t(\xB^*) - (\fbar(\xB_t) -\fbar(\xB^*))$, $t = 1, \ldots, T$.
We observe that $\EBB[s_t | \FM_{t - 1}] = 0$, where $\FM_{t - 1}$ is the $\sigma$-algebra generated by $\{f_1, \xi_1, \ldots, f_{t - 1}, \xi_{t-1}\}$.
This implies that $\{s_t\}_{t=1}^T$ is a martingale difference sequence.
By Assumption~\ref{assumption_stoch}.\ref{assumption_stoch_value}, we have
\begin{equation}
    |s_t| = \big| f_t(\xB_t) - f_t(\xB^*) - (\fbar(\xB_t) - \fbar(\xB^*)) \big| \le 2 M.
\end{equation}
Applying Proposition~\ref{proposition_concentration} to the sequence $s_t$, we obtain
\begin{equation}
    \PBB ( |\sum_{t=1}^T s_t| \ge \lambda) \le 4 \mathrm{exp}( - \frac{\lambda^2}{16 T M^2}),
\end{equation}
where $\lambda$ is an arbitrary positive number.
By setting $\lambda = 4 M \sqrt{T \mathrm{log}(8 / \delta)}$, we have w.p. at least $1 - \delta / 2$,
\begin{equation}
    \sum_{t=1}^T s_t
    = \sum_{t=1}^T \big( f_t(\xB_t) - f_t(\xB^*) \big) - \sum_{t=1}^T \big( \fbar(\xB_t) - \fbar(\xB^*) \big)
    \le 4 M \sqrt{T \mathrm{log}(8 / \delta)}.
\end{equation}
After rearranging terms, we get
\begin{equation}  \label{eq_reduction_whp}
    \RM_T = \sum_{t=1}^T \big( f_t(\xB_t) - f_t(\xB^*) \big)
    \le \sum_{t=1}^T \big( \fbar(\xB_t) - \fbar(\xB^*) \big) + 4 M \sqrt{T \mathrm{log}(8 / \delta)}.
\end{equation}
Hence, it remains to show that $\sum_{t=1}^T \big( \fbar(\xB_t) - \fbar(\xB^*) \big) = \tilde{\OM}(\sqrt{T})$.
By recursively applying Lemma~\ref{lemma_descent_convex}, we have $\forall t \ge 1$,
\begin{equation}  \label{eq_recursion_stochastic_whp}
\begin{aligned}[b]
    \fbar(\xB_t) - \fbar(\xB^*)
    &\le (1 - \eta_t) \big( \fbar(\xB_{t}) - \fbar(\xB^*) \big) + \eta_t D \| \epsilonB_t \| + \frac{L D^2 \eta_t^2}{2} \\
    &\le \prod_{\tau = 1}^{t - 1} (1 - \eta_{\tau}) \big( \fbar(\xB_1) - \fbar(\xB^*) \big)
        + \sum_{\tau = 1}^{t - 1} \eta_{\tau} \Big( D \|\epsilonB_{\tau}\| + \frac{L D^2 \eta_{\tau}}{2} \Big) \prod_{k = \tau + 1}^{t - 1} (1 - \eta_k)  \\
    &= \frac{1}{t} \big( \fbar(\xB_1) - \fbar(\xB^*) \big)
        + \sum_{\tau = 1}^{t - 1} \frac{1}{\tau + 1} \Big( D \|\epsilonB_{\tau}\| + \frac{L D^2}{2(\tau + 1)} \Big) \frac{\tau + 1}{t}  \\
    &= \frac{1}{t} \big( \fbar(\xB_1) - \fbar(\xB^*) \big)
        + \frac{1}{t} \sum_{\tau = 1}^{t - 1} \Big( D \|\epsilonB_{\tau}\| + \frac{L D^2}{2(\tau + 1)} \Big)  \\
    &\le \frac{1}{t} \big( \fbar(\xB_1) - \fbar(\xB^*) \big)
        + \frac{D}{t} \sum_{\tau = 1}^{t - 1} \|\epsilonB_{\tau}\|
        + \frac{ L D^2 \mathrm{log}t }{2 t},
\end{aligned}
\end{equation}
where the first equality follows from the choice of $\eta_{\tau}$.
Summing~\eqref{eq_recursion_stochastic_whp} from $t = 1$ to $T$, we obtain
\begin{equation} \label{eq_sum_recursion_stochastic_whp}
\begin{aligned}[b]
    \sum_{t = 1}^T \big( \fbar(\xB_t) - \fbar(\xB^*) \big)
    &\le \sum_{t=1}^T \frac{1}{t} \big( \fbar(\xB_1) - \fbar(\xB^*) \big)
        + \sum_{t=1}^T \sum_{\tau = 1}^{t - 1} \frac{D}{t} \|\epsilonB_{\tau}\|
        + \sum_{t=1}^T \frac{ L D^2 \mathrm{log}t }{2 t} \\
    &\le \sum_{t=1}^T \frac{1}{t} \big( \fbar(\xB_1) - \fbar(\xB^*) \big)
        + \sum_{t=1}^T \sum_{\tau = 1}^{t - 1} \frac{D}{t} \|\epsilonB_{\tau}\|
        + \frac{ L D^2 \mathrm{log}T }{2} \sum_{t=1}^T \frac{1}{t} \\
    &\le (\mathrm{log} T + 1) \big( \fbar(\xB_1) - \fbar(\xB^*) \big)
        + \sum_{t=1}^T \sum_{\tau = 1}^{t - 1} \frac{D}{t} \|\epsilonB_{\tau}\|
        + \frac{ L D^2 (\mathrm{log}T + 1)^2 }{2}.
\end{aligned}
\end{equation}
By Lemma~\ref{lemma_gradient_error_bound_whp} and the union bound, we have with probability at leat $1 - \delta / 2$,
\begin{equation}  \label{eq_union_bound_whp}
\begin{aligned}[b]
    \sum_{t=1}^T \sum_{\tau = 1}^{t - 1} \frac{D}{t} \|\epsilonB_{\tau}\|
    &\le 4 (L D^2 + \sigma D) \sqrt{2 \mathrm{log}(8 T / \delta)} \sum_{t=1}^T \sum_{\tau = 1}^{t - 1} \frac{1}{t \sqrt{\tau + 1}} \\
    &\le 4 (L D^2 + \sigma D) \sqrt{2 \mathrm{log}(8 T / \delta)} \sum_{t=1}^T \frac{2 \sqrt{t}}{t} \\
    &\le 16 (L D^2 + \sigma D) \sqrt{2 T \mathrm{log}(8 T / \delta)}.
\end{aligned}
\end{equation}
Plugging~\eqref{eq_union_bound_whp} into~\eqref{eq_sum_recursion_stochastic_whp}, we have with probability at leat $1 - \delta / 2$,
\begin{equation}  \label{eq_sum_union_whp}
    \sum_{t = 1}^T \big( \fbar(\xB_t) - \fbar(\xB^*) \big)
    \le (\mathrm{log} T + 1) \big( \fbar(\xB_1) - \fbar(\xB^*) \big)
        + \frac{ L D^2 (\mathrm{log}T + 1)^2 }{2}
        + 16 (L D^2 + \sigma D) \sqrt{2 T \mathrm{log}(8 T / \delta)}.
\end{equation}
Combining~\eqref{eq_reduction_whp} and~\eqref{eq_sum_union_whp} and applying the union bound, we have w.p. at least $1 - \delta$,
\begin{equation}
    \RM_T \le (\mathrm{log} T + 1) \big( \fbar(\xB_1) - \fbar(\xB^*) \big)
    + \frac{ L D^2 (\mathrm{log}T + 1)^2 }{2}
    + (16 L D^2 + 16 \sigma D + 4 M) \sqrt{2 T \mathrm{log}(8 T / \delta)}
    = \tilde{\OM}(\sqrt{T}),
\end{equation}
which is the desired result.
\end{proof}

\subsection{Proof of Corollary~\ref{corollary_convex_convergence}}
\begin{proof}
    Following the same argument as~\eqref{eq_sum_union_whp}, we have w.p. at least $1 - \delta$,
    \begin{equation}
        \frac{1}{T} \sum_{t = 1}^T \fbar(\xB_t) - \fbar(\xB^*)
        \le \frac{\mathrm{log}T + 1}{T} \big( \fbar(\xB_1) - \fbar(\xB^*) \big)
            + \frac{ L D^2 (\mathrm{log}T + 1)^2 }{2 T} \\
            + 16 (L D^2 + \sigma D) \frac{\sqrt{2 \mathrm{log}(4 T / \delta)}}{\sqrt{T}}.
    \end{equation}
    On the other hand, by the convexity of $\fbar$ and Jensen's inequality, we have
    \begin{equation}
        \fbar(\frac{1}{T} \sum_{t = 1}^T \xB_t)
        \le \frac{1}{T} \sum_{t = 1}^T \fbar(\xB_t).
    \end{equation}
    Combining the above two inequalities leads to the desired result.
\end{proof}

\subsection{Proof of Proposition~\ref{proposition_nonconvex_convergence}}
To begin with, we present the following useful lemma.
\begin{lemma} \label{lemma_descent_nonconvex}
    Consider the proposed ORGFW method.
    If the conditions in Assumptions~\ref{assumption_constraint} and \ref{assumption_stoch_smooth} are satisfied, then
    \begin{equation}
        \fbar(\xB_{t+1}) - \fbar(\xB_t) \le - \eta_t \GM(\xB_t) + 2 \eta_t D \| \epsilonB_t \| + \frac{L D^2 \eta_t^2}{2},
    \end{equation}
    where $\GM(\cdot)$ is the Frank-Wolfe gap, i.e., $\GM(\xB) = \max_{\uB \in \CM} \langle \nabla \fbar(\xB), \xB - \uB \rangle$.
\end{lemma}

\begin{proof}
    By Assumption~\ref{assumption_stoch_smooth}, we have
    \begin{equation} \label{eq_descent_lemma_start_point_nonconvex}
    \begin{aligned}[b]
        \fbar(\xB_{t+1}) - \fbar(\xB_t)
        &\le \langle \nabla \fbar(\xB_t), \xB_{t+1} - \xB_t \rangle
            + \frac{L}{2} \| \xB_{t+1} - \xB_t \|^2 \\
        &\le \eta_t \langle \nabla \fbar(\xB_t), \vB_{t} - \xB_t \rangle
            + \frac{\eta_t^2 L D^2}{2} \\
        &= \eta_t \langle \nabla \fbar(\xB_t) - \dB_t, \vB_{t} - \xB_t \rangle
            + \eta_t \langle \dB_t, \vB_{t} - \xB_t \rangle
            + \frac{\eta_t^2 L D^2}{2} \\
        &\le \eta_t D \| \epsilonB_t \|
            + \eta_t \langle \dB_t, \vB_{t} - \xB_t \rangle
            + \frac{\eta_t^2 L D^2}{2}.
    \end{aligned}
    \end{equation}
    Denote $\vB_t^+ = \mathrm{argmax}_{\vB \in \CM} \langle \nabla \fbar(\xB_t), \xB_t - \vB \rangle$, then
    \begin{equation} \label{eq_FW_gap_bound}
    \begin{aligned}[b]
        \GM(\xB_t)
        &= \langle \nabla \fbar(\xB_t), \xB_t - \vB_t^+ \rangle
        = \langle \nabla \fbar(\xB_t) - \dB_t, \xB_t - \vB_t^+ \rangle
            + \langle \dB_t, \xB_t - \vB_t^+ \rangle  \\
        & \le \| \epsilonB_t \| D + \langle \dB_t, \xB_t - \vB_t^+ \rangle
        \le \| \epsilonB_t \| D + \langle \dB_t, \xB_t - \vB_t \rangle,
    \end{aligned}
    \end{equation}
    where the first inequality follows from Cauchy-Schwarz and Assumption~\ref{assumption_constraint} and the second inequality follows from the fact that $\vB_t \in \mathrm{argmin}_{\xB \in \CM} \langle \dB_t, \vB \rangle$.
    Combining~\eqref{eq_descent_lemma_start_point_nonconvex} and \eqref{eq_FW_gap_bound} leads to
    \begin{equation}
    \begin{aligned}[b]
        \fbar(\xB_{t+1}) - \fbar(\xB_t)
        &\le \eta_t D \| \epsilonB_t \|
            + \eta_t D \| \epsilonB_t \| - \eta_t \GM(\xB_t)
            + \frac{\eta_t^2 L D^2}{2} \\
        &= - \eta_t \GM(\xB_t)
            + 2 \eta_t D \| \epsilonB_t \|
            + \frac{\eta_t^2 L D^2}{2},
    \end{aligned}
    \end{equation}
    which is the desired result.
\end{proof}

Now we are ready to prove Proposition~\ref{proposition_nonconvex_convergence}.

\begin{proof}
    (Proof of Proposition~\ref{proposition_nonconvex_convergence})
    By plugging $\rho_t = \eta_t = 1/(t + 1)^{2/3}$ into Lemma~\ref{lemma_gradient_error_bound_whp}, we have w.p. at least $1 - \delta / T$,
    \begin{equation}   \label{eq_gradient_error_sqr_nonconvex}
        \| \epsilonB_t \| \le 2 (2 L D + 3 \sigma)(t + 1)^{-1/3} \sqrt{2 \mathrm{log}(4 T / \delta)},
    \end{equation}
    where $\delta \in (0, 1)$ is a constant.
    Applying~\eqref{eq_gradient_error_sqr_nonconvex} to Lemma~\ref{lemma_descent_nonconvex}, we have w.p. at least $1 - \delta / T$,
    \begin{equation} \label{eq_descent_nonconvex}
    \begin{aligned}[b]
        \eta_t \GM(\xB_t)
        &\le \fbar(\xB_t) - \fbar(\xB_{t+1})
            + 2 (2 L D^2 + 3 \sigma^2)(t + 1)^{-1} \sqrt{2 \mathrm{log}(4 T / \delta)}
            + \frac{2 L D^2}{(t+1)^{4/3}} \\
        &\le \fbar(\xB_t) - \fbar(\xB_{t+1})
            + \frac{C_2'}{t+1},
    \end{aligned}
    \end{equation}
    where $C_2' = 2 (2 L D^2 + 3 \sigma^2) \sqrt{2 \mathrm{log}(4 T / \delta)} + 2 L D^2$.
    Summing~\eqref{eq_descent_nonconvex} from $t = 1$ to $T$ and using the union bound, we have w.p. at least $1 - \delta$,
    \begin{equation}
        \sum_{t=1}^T \eta_t \GM(\xB_t)
        \le \fbar(\xB_1) - \fbar(\xB^*) + C_2' \mathrm{log}(T + 1)
    \end{equation}
    Since $\eta_t \ge \eta_T = \frac{1}{(T+1)^{2/3}}$ for any $t \in \{ 1, \ldots, T \}$, we have w.p. at least $1 - \delta$,
    \begin{equation}
        \frac{1}{(T+1)^{2/3}} \sum_{t=1}^T \GM(\xB_t)
        \le \fbar(\xB_1) - \fbar(\xB^*) + C_2' \mathrm{log}(T + 1)
    \end{equation}
    To simplify notation, we denote
    $
        C_2 = \fbar(\xB_1) - \fbar(\xB^*)
                + C_2' \mathrm{log}(T + 1)
            = \fbar(\xB_1) - \fbar(\xB^*)
                + (2 \sqrt{2} (2 L D^2 + 3 \sigma^2) \sqrt{\mathrm{log}(4 T / \delta)} + 2 L D^2) \mathrm{log}(T + 1).
    $
    Rearranging terms, we have w.p. at least $1 - \delta$,
    \begin{equation}  \label{eq_FW_gap_final_nonconvex}
        \frac{1}{T} \sum_{t=1}^T \GM(\xB_t)
        \le \frac{1}{T} C_2 (T + 1)^{2/3}
        \le \frac{2 C_2}{T^{1/3}}.
    \end{equation}
    Substituting the LHS of~\eqref{eq_FW_gap_final_nonconvex} with $\min_{1 \le t \le T} \GM(\xB_t)$ leads to the desired result.
\end{proof}

\subsection{Proof of Theorem~\ref{theorem_adversarial}}
\begin{proof}
    By Assumption~\ref{assumption_adversarial_smooth}, we have
    \begin{equation} \label{eq_start_point_adversarial}
    \begin{aligned}[b]
        f_t(\xB_t^{(k+1)}) - f_t(\xB^*)
        &= f_t(\xB_t^{(k)} + \eta_k(\vB_t^{(k)} - \xB_t^{(k)})) - f_t(\xB^*) \\
        &\le f_t(\xB_t^{(k)}) - f_t(\xB^*)
            + \eta_k \langle \nabla f_t(\xB_t^{(k)}), \vB_t^{(k)} - \xB_t^{(k)} \rangle
            + \frac{L \eta_k^2}{2} \| \vB_t^{(k)} - \xB_t^{(k)} \|^2 \\
        &\le f_t(\xB_t^{(k)}) - f_t(\xB^*)
            + \eta_k \langle \nabla f_t(\xB_t^{(k)}), \vB_t^{(k)} - \xB_t^{(k)} \rangle
            + \frac{L \eta_k^2 D^2}{2}
    \end{aligned}
    \end{equation}
    where $\xB^* \in \mathrm{argmin}_{\xB \in \CM} \sum_{t=1}^T f_t(\xB)$ and the last inequality follows from Assumption~\ref{assumption_constraint}.
    We observe that
    \begin{equation} \label{eq_inner_product_bound_adversarial}
    \begin{aligned}[b]
        \langle \nabla f_t(\xB_t^{(k)}), \vB_t^{(k)} - \xB_t^{(k)} \rangle
        &= \langle \nabla f_t(\xB_t^{(k)}) - \dB_t^{(k)}, \vB_t^{(k)} - \xB_t^{(k)} \rangle
            + \langle \dB_t^{(k)}, \vB_t^{(k)} - \xB_t^{(k)} \rangle \\
        &= \langle \nabla f_t(\xB_t^{(k)}) - \dB_t^{(k)}, \vB_t^{(k)} - \xB_t^{(k)} \rangle
            + \langle \dB_t^{(k)}, \vB_t^{(k)} - \xB^* \rangle
            + \langle \dB_t^{(k)}, \xB^* - \xB_t^{(k)} \rangle \\
        &= \langle \nabla f_t(\xB_t^{(k)}) - \dB_t^{(k)}, \vB_t^{(k)} - \xB^* \rangle
            + \langle \dB_t^{(k)}, \vB_t^{(k)} - \xB^* \rangle
            + \langle \nabla f_t(\xB_t^{(k)}), \xB^* - \xB_t^{(k)} \rangle \\
        &\le \langle \nabla f_t(\xB_t^{(k)}) - \dB_t^{(k)}, \vB_t^{(k)} - \xB^* \rangle
            + \langle \dB_t^{(k)}, \vB_t^{(k)} - \xB^* \rangle
            + f_t(\xB^*) - f_t(\xB_t^{(k)}) \\
        &\le D \cdot \| \nabla f_t(\xB_t^{(k)}) - \dB_t^{(k)} \|
            + \langle \dB_t^{(k)}, \vB_t^{(k)} - \xB^* \rangle
            + f_t(\xB^*) - f_t(\xB_t^{(k)})
    \end{aligned}
    \end{equation}
    where the first inequality follows from convexity of $f_t$ and the second inequality follows from the Cauchy-Schwarz inequality and Assumption~\ref{assumption_constraint}.
    Plugging \eqref{eq_inner_product_bound_adversarial} into \eqref{eq_start_point_adversarial} leads to
    \begin{equation} \label{eq_recursion_adversarial}
    \begin{aligned}[b]
        f_t(\xB_t^{k+1}) - f_t(\xB^*)
        &\le (1 - \eta_k) \big( f_t(\xB_t^{(k)}) - f_t(\xB^*) \big)
            + \eta_k D \| \nabla f_t(\xB_t^{(k)}) - \dB_t^{(k)} \|
            + \eta_k \langle \dB_t^{(k)}, \vB_t^{(k)} - \xB^* \rangle
            + \frac{L \eta_k^2 D^2}{2}
    \end{aligned}
    \end{equation}
    To simplify notation, we denote $\epsilonB_t^{(k)} := \nabla f_t(\xB_t^{(k)}) - \dB_t^{(k)}$.
    Following the same argument as Lemma~\ref{lemma_gradient_error_bound_whp}, if we set $\rho_k = \eta_k = 1/(k+1)$, we have w.p. at least $1 - \delta / (T K)$,
    \begin{equation} \label{eq_gradient_error_sqr_adversarial}
        \|\epsilonB_t^{(k)} \| \le \frac{C_3}{\sqrt{k + 1}},
    \end{equation}
    where $\delta \in (0, 1)$ and $C_3 = 4 (L D + \hat{\sigma}) \sqrt{2 \mathrm{log}(4 T K / \delta)}$.
    Combining~\eqref{eq_recursion_adversarial} and~\eqref{eq_gradient_error_sqr_adversarial}, we have w.p. at least $1 - \delta / (T K)$
    \begin{equation}
    \begin{aligned}[b]
        f_t(\xB_t^{(k+1)}) - f_t(\xB^*)
        &\le (1 - \eta_k) \big( f_t(\xB_t^{(k)}) - f_t(\xB^*) \big)
            + \eta_k \Big( \frac{C_3}{\sqrt{k + 1}}
            + \frac{\eta_k L D^2}{2}
            + \langle \dB_t^{(k)}, \vB_t^{(k)} - \xB^* \rangle \Big)
    \end{aligned}
    \end{equation}
    Summing the above inequality from $t = 1$ to $T$ and applying the union bound, we have w.p. at least $1 - \delta / K$,
    \begin{equation} \label{eq_sum_t_adversarial}
    \begin{aligned}[b]
        \sum_{t=1}^T \Big( f_t(\xB_t^{(k+1)}) - f_t(\xB^*) \Big)
        &\le (1 - \eta_k) \sum_{t=1}^T \Big( f_t(\xB_t^{(k)}) - f_t(\xB^*) \Big) \\
        & \ \ \ \
            + \eta_k \Big( \frac{C_3 T}{\sqrt{k + 1}}
            + \frac{\eta_k L D^2 T}{2}
            + \sum_{t=1}^T \langle \dB_t^{(k)}, \vB_t^{(k)} - \xB^* \rangle \Big)
    \end{aligned}
    \end{equation}
    We observe that the last term of~\eqref{eq_sum_t_adversarial} can be bounded from above as follows
    \begin{equation} \label{eq_olo_reget_adversarial}
        \sum_{t=1}^T \langle \dB_t^{(k)}, \vB_t^{(k)} - \xB^* \rangle
        \le \sum_{t=1}^T \langle \dB_t^{(k)}, \vB_t^{(k)} \rangle  - \min_{\xB \in \CM}\sum_{t=1}^T \langle \dB_t^{(k)}, \xB \rangle
        = \RM_T^{\EM},
    \end{equation}
    where $\RM_T^{\EM}$ is the regret of the OLO algorithms $\EM^{(1)}, \ldots, \EM^{(K)}$.
    To simplify notation, we denote $\Psi_k = \sum_{t=1}^T \Big( f_t(\xB_t^{(k)}) - f_t(\xB^*) \Big)$.
    Plugging~\eqref{eq_olo_reget_adversarial} into~\eqref{eq_sum_t_adversarial} and setting $k = K$, we have w.p. at least $1 - \delta$,
    \begin{equation} \label{eq_sum_k_adversarial}
    \begin{aligned}[b]
        \Psi_{K+1}
        &\le (1 - \eta_K) \Psi_K
            + \eta_K \Big( \frac{C_3 D T}{\sqrt{K+1}} + \frac{\eta_K L D^2 T}{2} + \RM_T^{\EM} \Big) \\
        &\le \prod_{k=1}^K (1 - \eta_k) \Psi_1
            + \sum_{k=1}^K \eta_k \prod_{j=k+1}^K (1 - \eta_j) \Big( \frac{C_3 D T}{\sqrt{k+1}} + \frac{\eta_k L D^2 T}{2} + \RM_T^{\EM} \Big).
    \end{aligned}
    \end{equation}
    Since we set $\eta_k = 1 / (k + 1)$, we have
    \begin{equation} \label{eq_product_expension_adversarial}
        \prod_{j=r}^K (1 - \eta_j) = \frac{K}{K+1} \cdot \frac{K-1}{K} \cdot \ldots \cdot \frac{r+1}{r+2} \cdot \frac{r}{r+1} = \frac{r}{(K + 1)}, \; \forall 1 \le r \le K.
    \end{equation}
    Plugging~\eqref{eq_product_expension_adversarial} into~\eqref{eq_sum_k_adversarial} and setting $K = T$, we have w.p. at least $1 - \delta$,
    \begin{equation}
    \begin{aligned}[b]
        \Psi_{K+1}
        &\le \frac{\Psi_1}{K + 1}
            + \sum_{k=1}^K \frac{1}{k+1} \cdot \frac{k}{K + 1} \Big( \frac{C_3 D T}{\sqrt{k+1}} + \frac{L D^2 T}{k+1} + \RM_T^{\EM} \Big) \\
        &\le \frac{\Psi_1}{K + 1}
            + \frac{1}{(K+1)} \sum_{k=1}^K \Big( \frac{C_3 D T}{\sqrt{k+1}} + \frac{L D^2 T}{k+1} + \RM_T^{\EM} \Big) \\
        &\le \frac{\Psi_1}{K + 1}
            + \frac{4 C_3 D T}{\sqrt{K+1}} + \frac{2 L D^2 T \mathrm{log}(K+1)}{K+1} + \RM_T^{\EM} \\
        &\le \frac{1}{T + 1} \sum_{t=1}^T \Big( f_t(\xB_1) - f_t(\xB^*) \Big)
            + 4 C_3 D \sqrt{T} + 2 L D^2 \mathrm{log}(T+1) + \RM_T^{\EM} \\
        &\le Q
            + 4 C_3 D \sqrt{T} + 2 L D^2 \mathrm{log}(T+1) + \RM_T^{\EM},
    \end{aligned}
    \end{equation}
    where $Q = \max_{1 \le t \le T} \{f_t(\xB_1) - f_t(\xB^*)\}$.
    By recalling that $\xB_t = \xB_t^{(K+1)}$, we have w.p. at least $1 - \delta$,
    \begin{equation}
    \begin{aligned}[b]
        \RM_T
        &= \sum_{t=1}^T f_t(\xB_t) - \sum_{t=1}^T f_t(\xB^*)
        = \Psi_{K+1} \\
        &\le 16 (L D^2 + \hat{\sigma} D) \sqrt{ T \mathrm{log}(4 T^2 / \delta)}
            + 2 L D^2 \mathrm{log}(T+1) + Q + \RM_T^{\EM}.
    \end{aligned}
    \end{equation}
    This completes the proof.
\end{proof}

\end{document}